\documentclass[sigconf]{acmart}
\copyrightyear{2026}
\acmYear{2026}
\setcopyright{cc}
\setcctype{by}
\acmConference[MM '26]{Proceedings of the 34th ACM International Conference on Multimedia}{November 10--14, 2026}{Rio de Janeiro, Brazil}
\acmBooktitle{Proceedings of the 34th ACM International Conference on Multimedia (MM '26), November 10--14, 2026, Rio de Janeiro, Brazil}
\acmDOI{10.1145/3767308.3835367}
\acmISBN{979-8-4007-2213-4/2026/11}
\usepackage{mathtools}
\usepackage{amsmath}
\usepackage[table,xcdraw]{xcolor}
\usepackage{multirow}
\usepackage{colortbl}

\begin{document}

\title{OASIS: Occlusion-aware Single-image Hand Avatar Reconstruction via 3D Gaussian Splatting}

\author{Zhisheng Han}
\affiliation{%
  \institution{University of Leicester}
  \city{Leicester}
  \country{UK}
}
\email{zh174@leicester.ac.uk}

\author{Shiyao Wu}
\affiliation{%
  \institution{University of Leicester}
  \city{Leicester}
  \country{UK}
}
\email{sw658@leicester.ac.uk}

\author{Jiayan Qiu}
\affiliation{%
  \institution{University of Leicester}
  \city{Leicester}
  \country{UK}
}
\email{jiayan.qiu.1991@outlook.com}

\author{Yakun Ju}
\affiliation{%
  \institution{University of Leicester}
  \city{Leicester}
  \country{UK}
}
\email{kelvin.yakun.ju@gmail.com}

\author{Lu Liu}
\affiliation{%
  \institution{University of Exeter}
  \city{Exeter}
  \country{UK}}
\email{l.liu3@exeter.ac.uk}

\author{Le Zhang}
\affiliation{%
  \institution{University of Birmingham}
  \city{Birminghan}
  \country{UK}}
\email{l.zhang.16@bham.ac.uk}

\author{Pengfei Feng}
\affiliation{%
  \institution{China University of Geoscience}
  \city{Wuhan}
  \country{China}}
\email{20161004188@cug.edu.cn}

\author{Huiyu Zhou}
\affiliation{%
  \institution{University of Leicester}
  \city{Leicester}
  \country{UK}
}
\email{hz143@leicester.ac.uk}

\author{Zheheng Jiang}
\correspondingauthor
\affiliation{%
  \institution{University of Leicester}
  \city{Leicester}
  \country{UK}
}
\email{zhehengjiangcareer@gmail.com}

\renewcommand{\shortauthors}{Zhisheng Han et al.}

\begin{abstract}
Single-image 3D hand avatar reconstruction is fundamentally ill-posed and particularly challenging due to limited visual evidence under severe self-occlusion and the complex pose-dependent deformation of highly articulated hands. Existing methods predominantly rely on implicit NeRF-style representations, whose volumetric fitting is computationally expensive and often struggles to preserve fine-grained hand details. In this work, we present OASIS, a tailored 3D Gaussian Splatting framework for single-image hand avatar reconstruction. To faithfully encode sparse image-specific appearance cues in single-view reconstruction, we construct geometry-aligned visual evidence tokens by explicitly aligning input image observations with 3D hand geometry and context-adaptively tokenizing the resulting visual evidence. Since severe self-occlusion makes the reliability of image evidence inherently visibility-dependent, we introduce a visibility-conditioned point-image attention to reliably transfer visual evidence to geometric tokens, yielding occlusion-aware Gaussian features for faithful and robust reconstruction. To further capture non-rigid deformation of articulated hands, we introduce a Feature-on-Mesh representation to enable Gaussian deformation to be guided by local surface stretching. Under this framework, we adopt a one-shot adaptation scheme that learns a shared hand prior from multi-identity training data and then fits it to a target image for target-specific reconstruction. Extensive experiments show that OASIS outperforms existing baselines in both visual fidelity and efficiency across challenging poses and in-the-wild scenarios, and further demonstrates strong versatility in downstream applications such as text-to-avatar generation and texture editing. Code will be released on the \href{https://mova-hand.github.io/MOVA/}{\textcolor{magenta}{project page}}.
\end{abstract}



\begin{teaserfigure}
  \includegraphics[width=\textwidth]{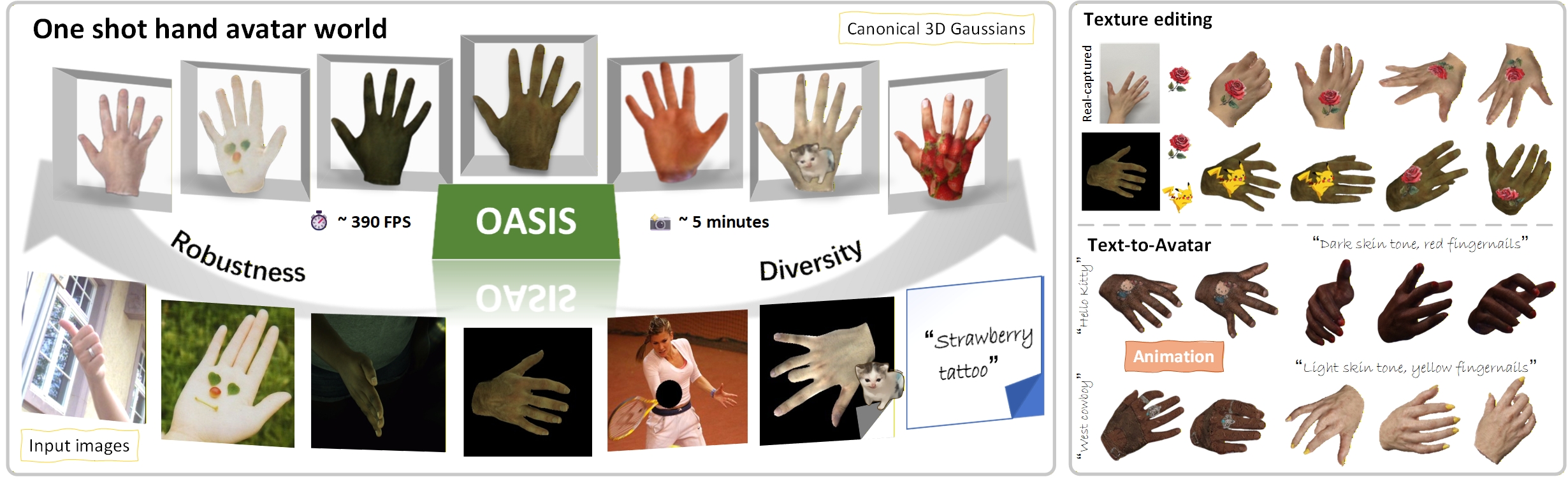}
  \caption{We introduce \textit{OASIS}, a 3DGS-based one-shot hand avatar reconstruction model with efficient adaptation ($\sim$5 minutes) and real-time rendering ($\sim$390 FPS). We showcase a gallery of one-shot hand avatars that highlights both robustness and diversity: it reconstructs high-fidelity hands under challenging poses, complex textures, and extreme side-view observations, while also supporting human-image inputs and downstream applications such as texture editing and text-to-avatar generation.}
  \Description{}
  \label{fig:teaser}
\end{teaserfigure}


\maketitle

\section{Introduction}
Single-image 3D avatar reconstruction is an emerging modeling technique that aims to create a life-like 3D avatar from just one image, offering a highly cost-effective and promising approach to 3D modeling \cite{qiu2025anigs, lhm}. Recent advances have made this direction increasingly attractive for realistic digital human modeling \cite{chen2024generalizable,lhm,zhang2025guava,wu2025sings,qiu2025anigs}. Among different avatar types, hand avatars are of particular importance because hands are a primary medium for interacting with the physical world through grasping and manipulation, while also providing rich expressive cues for gesture-based communication in virtual environments \cite{kim2024bitt,tu2023consistent,Moon_2024_CVPR,dong2024hamba}.

Despite its practical importance, hand avatar reconstruction remains relatively underexplored compared with full-body avatar modeling. Prior hand avatar methods \cite{corona2022lisa, handavatar, handnerf} have largely relied on Neural Radiance Fields (NeRF) \cite{nerf} and related implicit neural fields. However, implicit volumetric rendering remains a practical bottleneck for avatar applications, as dense ray sampling and repeated field evaluations are computationally expensive, limiting interactivity and practical deployment \cite{qian20243dgsavatar, splattingavatar}. To address this broader limitation of implicit avatar rendering, recent human research \cite{zhang2024mesh,hu2024gauhuman, lhm, zhang2025guava,qiu2025anigs,wu2025sings} has increasingly adopted explicit 3D Gaussian Splatting (3DGS) \cite{3dgs} representation, which enables efficient rendering while maintaining photorealistic quality. Nevertheless, exploiting these advantages in hand avatar reconstruction remains non-trivial.

Recent 3DGS-based avatar methods achieve high reconstruction fidelity and rendering efficiency for animatable human avatars. However, extending them to single-view hand reconstruction is non-trivial, as hands exhibit significantly greater articulation and self-occlusion than body or head avatars, making single-image visual evidence sparse and ambiguous. To address this, we construct geometry-aligned visual evidence tokens that preserve image-specific appearance by aligning it with 3D hand geometry and encoding point-aligned observations into compact tokens. To address the challenge that self-occlusion deprives occluded points of reliable observations and makes them more susceptible to interference from other surfaces, we propose a visibility-conditioned point–image attention module that adaptively modulates the integration of image appearance cues into geometric tokens based on visibility. To further improve non-rigid hand modeling by capturing local surface stretching, we introduce a Feature-on-Mesh representation that organizes fused features over mesh faces. This design enables the decoded Gaussians to deform coherently with mesh-face stretching under articulation, yielding smoother and more consistent appearance.

Through training on accessible multi-identity data, the proposed single-image reconstruction framework learns prior reconstruction knowledge. As discussed in \cite{ohta}, one-shot hand avatar reconstruction requires target-specific personalization to adapt a prior model to the input image—capturing observed appearance while relying on the prior for unseen regions. Our goal is to inject identity-specific appearance and fine texture details into the Gaussian representation while preserving priors in unobserved areas. To this end, we first apply color calibration to compensate for global appearance discrepancies between the input image and the prior. We then fine-tune the Gaussian decoder with LoRA \cite{hu2022lora} to capture target-specific textures. This personalization enables robust identity adaptation and high-fidelity reconstruction for one-shot hand avatars. Our main contributions are:
\begin{itemize}

    \item 
    We introduce OASIS, a 3DGS-based one-shot hand avatar reconstruction model for reconstructing animatable and high-fidelity hand avatars from a single image.
    \item
    In OASIS, we first construct geometry-aligned visual evidence tokens and perform visibility-conditioned evidence transfer to produce occlusion-aware Gaussian features, enabling faithful and robust single-view hand reconstruction. We further introduce a Feature-on-Mesh representation to better capture local surface deformation under non-rigid hand articulation.
    
    \item 
    Extensive experiments on InterHand2.6M and in-the-wild images demonstrate that our method consistently outperforms existing approaches in fidelity, robustness, and rendering efficiency.
    We further show the versatility of our pipeline on text-to-avatar generation and texture editing.

\end{itemize}

\section{Related Work}

\noindent \textbf{Single-image Animatable Gaussian Avatar Reconstruction.} 
Reconstructing an animatable 3D avatar from a single image is fundamentally ill-posed due to incomplete observations and ambiguous geometry. Recent 3DGS-based methods alleviate this difficulty mainly through prior-guided synthesized observations and prior-driven reconstruction models. SinGS \cite{wu2025sings} leverages Kinematic Human Diffusion to synthesize pose-space auxiliary observations with high 3D consistency for single-image reconstruction, and further combines compact 3D distillation to consolidate incomplete monocular evidence into Gaussian avatars. AniGS \cite{qiu2025anigs} synthesizes detailed multi-view canonical-pose observations with a pretrained generative model to reduce reconstruction ambiguity, and addresses their cross-view inconsistency through a 4D Gaussian formulation. LHM \cite{lhm} adopts a large multimodal reconstruction model that fuses image and geometric features for direct Gaussian avatar prediction. LAM \cite{he2025lam} further exploits a canonical-space parametric prior to enable direct animatable Gaussian reconstruction in the head domain. Among these works, our approach is most closely related to prior-driven reconstruction models, which are better suitable for hands as they directly predict canonical-space Gaussians. By contrast, methods relying on auxiliary observations are less practical, as the strong hand articulation make such observations harder to synthesize consistently and accurately. Distinct from the human body and head, the hand is a highly articulated structure \cite{handavatar}, making single-image reconstruction more challenging due to severe self-occlusion, ambiguous monocular evidence, and fine-grained non-rigid deformation. These hand-specific challenges are not fully captured by existing 3DGS avatar methods. To address this gap, we propose a hand-tailored one-shot 3DGS reconstruction model.

\noindent \textbf{Animatable Hand Avatar.}
Existing research on animatable hand avatars mainly follows two lines. Early methods are predominantly based on explicit parametric hand models with texture representations. HTML \cite{qian2020html} introduces a parametric texture space on top of MANO model \cite{mano}. Building on this line, NIMBLE \cite{li2022nimble} and Handy \cite{potamias2023handy} further improve surface and texture realism through richer hand modeling and learned texture priors. HARP \cite{harp} advances this direction by adapting both geometry and appearance to better capture identity-specific characteristics. Nevertheless, their visual fidelity is still limited by the coarse geometry and sparsity of parametric hand meshes. More broadly, advances in neural avatars, such as \cite{weng2022humannerf} and \cite{jiang2022selfrecon}, have demonstrated the potential of neural representations for modeling articulated appearance. However, directly extending such full-body formulations to hands remains challenging due to the more severe self-contact, self-occlusion, and fine-grained appearance variations, motivating hand-specific methods for realistic reconstruction. LISA \cite{corona2022lisa} is the first to model hand shape and appearance with NeRF, while HandAvatar \cite{handavatar} further disentangles hand geometry, albedo, and illumination for high-fidelity animatable hand avatars. Subsequent methods such as HandNeRF \cite{handnerf} and LiveHand \cite{mundra2023livehand} improve deformation-aware neural rendering and real-time performance under hand motion. However, these methods typically rely on multi-view images or monocular videos. OHTA \cite{ohta} begins to explore one-shot hand avatar reconstruction from a single image by learning transferable hand priors and performing test-time inversion and fitting. However, it still builds on implicit volumetric representations, which are computationally expensive to optimize and render, and shows limited ability to faithfully reconstruct high-frequency details due to the indirect coupling between image observations and 3D structure, thereby limiting practical usability. In contrast, our work develops a 3DGS-based one-shot hand avatar framework, enabling efficient rendering together with faithful reconstruction from a single image.

\section{Method}
\subsection{Overview}
\begin{figure*}[t]
\begin{center}
\setlength{\belowcaptionskip}{0.6cm}
   \includegraphics[width=1\linewidth]{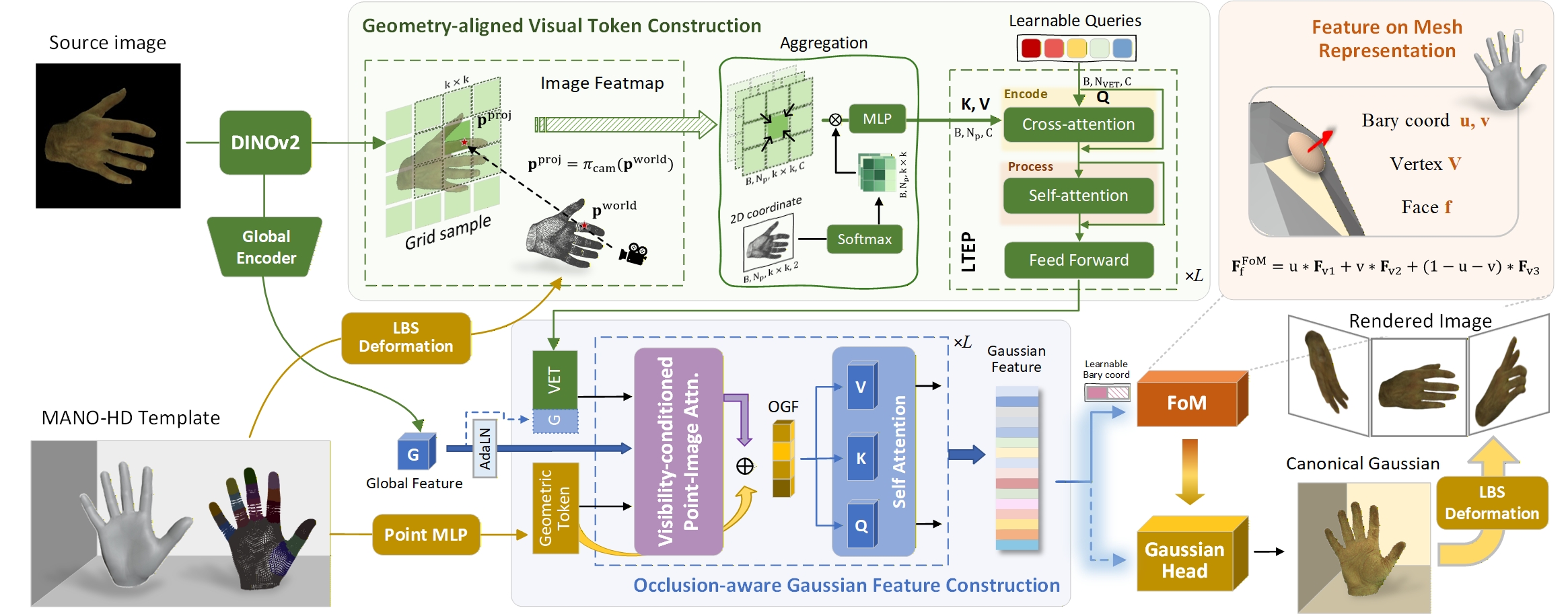}
   \captionsetup{font=small} 
   \caption {
   Overview of the proposed OASIS. Given a single input image, we construct geometry-aligned VETs to encode image-specific appearance cues, and use visibility-conditioned point-image cross-attention to transfer visual evidence to geometry-anchored point tokens, yielding occlusion-aware Gaussian features (OGF) under self-occlusion. We then refine these features with self-attention and lift them onto mesh faces via the FoM representation, providing a face-level carrier for Gaussian deformation under non-rigid hand animation. The final Gaussian features are decoded into Gaussian attributes in canonical space, followed by animating to target poses via LBS.
   } \label{fig:main}
\end{center}
\vspace{-20pt}
\end{figure*}

Given a single RGB hand image $\mathcal{I} \in \mathbb{R}^{H\times W\times 3}$, our goal is to reconstruct an animatable 3D hand avatar represented by 3D Gaussians. To this end, we propose OASIS, an animatable Gaussian hand reconstruction model, as shown in fig. \ref{fig:main}. Our method addresses three key challenges in single-image hand reconstruction: preserving sparse image-specific appearance evidence, robustly transferring such evidence to 3D representations under severe self-occlusion, and modeling local surface Gaussian deformation under hand articulation. Specifically, we design a geometry-aligned visual token construction module (Sec.~\ref{sec:VET}) that extracts geometry-aligned image evidence and encodes it into a compact set of visual evidence tokens (VETs). Built upon the geometry-aligned VETs, we further develop a visibility-conditioned point-image attention (Sec.~\ref{sec:occlusion}) module to reliably transfer visual evidence to geometric tokens under severe self-occlusion, producing occlusion-aware Gaussian features for robust hand reconstruction. Moreover, we introduce a Feature-on-Mesh (FoM) representation (Sec.~\ref{sec:fom}) over mesh faces, enabling Gaussian deformation to be guided by local stretching. Finally, we decode the fused features into a canonical set of per-Gaussian attributes, including position, scaling, rotation, opacity, and color, and animate them to target poses via linear blend skinning (LBS).

\subsection{Geometry-aligned Visual Token Construction}
\label{sec:VET}
In single-image hand avatar reconstruction, a central challenge lies in effectively lifting 2D appearance observations into 3D space for animatable representation. Prior work \cite{ohta} relies on an implicit volumetric paradigm, which is computationally expensive and only indirectly propagates 2D image evidence to 3D structure, making fine-grained and robust reconstruction difficult under severe self-occlusion and articulation. Recent human works \cite{lhm,zhang2025guava,wu2025fastavatar} instead adopt an explicit 3DGS reconstruction pipeline, typically using backbone image features and transferring them to geometric tokens. However, such generic image-feature transfer is insufficient for hands, where limited single-view evidence is further degraded by severe self-occlusion, making reliable association between local image observations and the 3D hand geometry substantially more difficult. Therefore, we introduce a geometry-guided projection to explicitly establish geometry-consistent correspondence, which aligns backbone image features with posed 3D points, converting image-domain representations into point-aligned visual observations. Building on these projected image features, we further propose a learnable tokenization scheme that content-adaptively encodes them into informative yet compact visual evidence tokens, reducing redundancy caused by nearby projections.

\subsubsection{Geometry-aligned Feature Projection and aggregation}
Given a source image, we extract dense image features using a DINOv2 backbone \cite{dinov2}, obtaining a feature map defined in the image domain $\mathbf{F}\in \mathbb{R}^{h\times w, C}$, where $h\times w$ denotes the spatial resolution of the feature map and $C$ is the feature dimension. In parallel, we make the initialization of 3D points $\{\mathbf{p}_i \in \mathbb{R}^3\}_{i=1}^{N_{\text{point}}}$ relying on the hand prior model MANO-HD \cite{handavatar}, a super-resolution version of hand model MANO \cite{mano}, to maintain the hand topology. Given the MANO shape and pose parameters estimated from the input image using an off-the-shelf estimator \cite{wilor}, these points could be deformed to posed space via LBS, resulting in posed 3D points. Instead of directly using backbone image features as in \cite{he2025lam,lhm,wu2025fastavatar}, we propose to align image features with posed hand geometry to make explicit 2D-3D correspondence for faithful reconstruction under limited visual cues. Specifically, we first project each posed point onto the image plane under the given camera extrinsic matrices: $\mathbf{p}_i^{\text{cam}} = \mathbf{R}\,\mathbf{p}_i + \mathbf{T}$, where $\mathbf{R} \in \mathbb{R}^{3 \times 3}$ and $\mathbf{T} \in \mathbb{R}^{3}$ denote the camera rotation and translation, respectively. The camera-space points are then projected onto the image plane using the camera intrinsic matrix $\mathbf{K}$: $\tilde{\mathbf{p}}_i = \mathbf{K}\,\mathbf{p}_i^{\text{cam}}$, followed by perspective division:
\begin{equation}
\setlength{\abovedisplayskip}{5pt} 
\setlength{\belowdisplayskip}{5pt}
\mathbf{p}_i^{\text{proj}} =
\left(
\frac{\tilde{\mathbf{p}}_{i,x}}{\tilde{\mathbf{p}}_{i,z}},
\frac{\tilde{\mathbf{p}}_{i,y}}{\tilde{\mathbf{p}}_{i,z}}
\right),
\end{equation}
where $\mathbf{p}_i^{\text{proj}} \in \mathbb{R}^2$ denotes the 2D pixel coordinate with image size $(H, W)$ corresponding to the $i$-th 3D point. Subsequently, we extract local image evidence from the feature maps by mapping projected pixel coordinates to the corresponding feature map with $h\times w$ resolution. Rather than relying on a single projection-aligned feature that is sensitive to projection errors, we adopt bilinear grid sampling on the image feature map over a local neighborhood around each projected position, yielding $K*K$ (with K=5) candidate visual evidence features that capture local appearance variation.

After obtaining the local image features aligned with each 3D point, we aggregate the sampled features into a single point-level image feature using a geometry-aware weighted pooling. For the $i$-th 3D point, let $\{ \mathbf{f}_{i,k} \in \mathbb{R}^C \}_{k=1}^{K}$ denote the set of local image features sampled from the feature map, with corresponding normalized coordinates $\{ \mathbf{c}_{i,k} \in \mathbb{R}^2 \}_{k=1}^{K}$. We compute a normalized 2D center for each 3D point $\bar{\mathbf{c}}_i$ from the sampled coordinates, and measure the relative distance between each sample and this center: $d_{i,k} = \left\lVert \mathbf{c}_{i,k} - \bar{\mathbf{c}}_i \right\rVert_2$. These distances are converted into spatial weights using a distance-based weighting function, assigning higher importance to the sampled features that are spatially closer to the projected point:
\begin{equation}
\setlength{\abovedisplayskip}{15pt} 
\setlength{\belowdisplayskip}{5pt}
w_{i,k} = \frac{\exp(-d_{i,k}/\sigma)}{\sum_{k'} \exp(-d_{i,k'}/\sigma)}.
\end{equation}
Therefore, the final point-aligned image feature is obtained as a weighted sum of the local image features: $\mathbf{f}^{\text{pa}}_i=\begin{matrix} \sum_{k=1}^{K}w_{i,k}\mathbf{f}_{i,k} \end{matrix}$.

\subsubsection{Learnable Tokenization via Encoding and Processing}
Although geometry-aligned projection establishes explicit correspondence between image observations and 3D points, the resulting $N_{\text{point}}$ point-aligned features $\mathbf{f}^{\text{pa}} \in \mathbb{R}^{N_{\text{point}} \times C}$ remain highly overcomplete. Due to redundant points and the limited spatial resolution of image features, many neighboring points receive nearly identical visual evidence. Consequently, directly treating point-aligned features as image tokens leads to representations dominated by redundant visual evidence rather than by informative visual content. To address this issue, we introduce a learnable tokenization module that compresses point-aligned features into a compact set of latent VETs. Specifically, we introduce a set of learnable query tokens $\mathbf{Q}_l$ to make tokenization a data-driven aggregation process. The learnable queries then enable the model to adaptively summarize the most informative and complementary point-aligned visual evidence into a fixed number of tokens through cross-attention:
\begin{equation}
\setlength{\abovedisplayskip}{5pt}
\setlength{\belowdisplayskip}{5pt}
\mathrm{Attn}(\mathbf{Q}_l, \mathbf{f}^{\text{pa}})
=
\mathrm{softmax}\!\left(
{\mathbf{Q}_l\mathbf{f}^{\text{pa}}}^\top / {\sqrt{d}} \right)\mathbf{f}^{\text{pa}},
\end{equation}
where $\mathbf{Q}_l \in \mathbb{R}^{N_{\text{vet}},C}$ serves as the query and $\mathbf{f}_{\text{pa}} \in \mathbb{R}^{N_{\text{point}},C}$ as the key and value. This encoding step adaptively aggregates the point-level features into $N_{\text{vet}}$ token-level representations, effectively enabling the model to allocate representational capacity based on visual content. While this stage captures point-level evidence independently for each token, it does not explicitly promote coordination among tokens. To address this, we introduce a subsequent processing stage based on self-attention, allowing tokens to interact and exchange information. Finally, a feed-forward network is applied to further refine its representation, yielding a compact and content-adaptive set of VETs, noted as $\mathbf{T}_\text{VET}$, for subsequent modeling.

\subsection{Occlusion-aware Gaussian Feature Construction}
\label{sec:occlusion}

\begin{figure}[t]
\begin{center}
\setlength{\belowcaptionskip}{0.5cm}
   \includegraphics[width=1\linewidth]{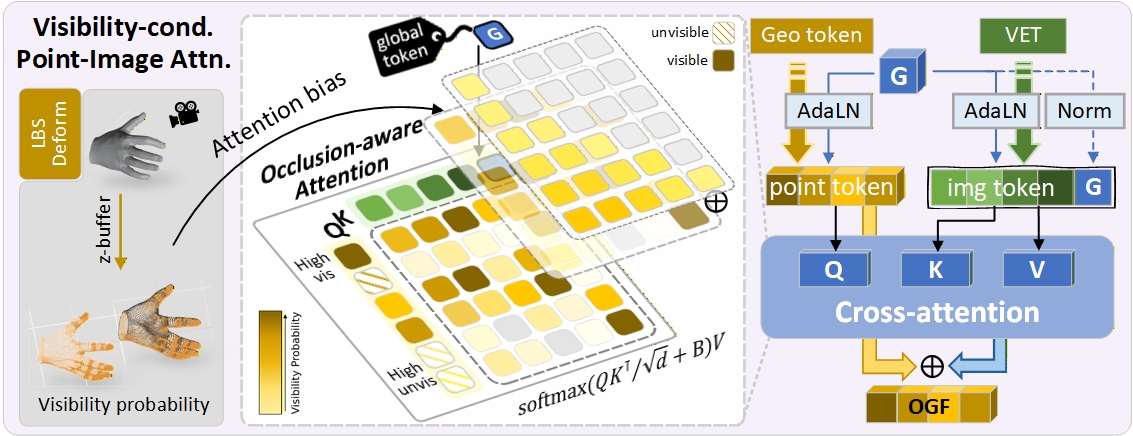}
   \captionsetup{font=small} 
   \caption {Architecture of the proposed Visibility-conditioned Point-Image Attention. OGF is short for occlusion-aware Gaussian features. } \label{fig:transformer}
\end{center}
\vspace{-30pt}
\end{figure}

After constructing geometry-aligned VETs, our next goal is to reliably transfer their visual evidence to geometry-anchored point tokens for point-wise appearance inference. Prior avatar methods model 3D-2D interaction either through cross-attention \cite{chen2024generalizable,chen2025synchuman,prospero2025gst} or unified multimodal transformers with full attention over geometric and visual tokens \cite{lhm}. However, under severe hand self-occlusion, occluded points project onto visible foreground regions rather than their own surface. In this case, feature-affinity attention cannot determine whether the matched image evidence is reliable for the queried point, making mismatched point-image associations difficult to suppress and causing erroneous appearance attribution. To address the above issues, we introduce visibility-conditioned point–image attention (VPIA), which explicitly modulates how each 3D point attends to image evidence based on its estimated visibility. This design encourages visible points to focus on reliable local evidence, while allowing occluded points to rely more on global visual context. To explicitly model dependencies among Gaussian features, we apply self-attention to propagate contextual information across features, enabling more coherent and robust reconstruction.

\subsubsection{Multimodal Token Representation}
Alongside the VETs, we introduce point-based geometric tokens that provide the structural foundation for subsequent visibility-conditioned point–image interactions and joint reasoning. The geometric tokens are derived from MANO-HD surface points and encode structural priors of the hand. By initialization on the MANO-HD template, each point is embedded via sinusoidal positional encoding \cite{nerf} followed by an multi-layer perceptron (MLP) projection: $\mathbf{T}_{3D} = \text{MLP}_{\text{proj}} (\gamma(\mathbf{p})) \in \mathbb{R}^{N_{\text{points}} \times C}$, where $\gamma : \mathbb{R}^3 \to \mathbb{R}^{3L}$ applies $L$-frequency sinusoidal encoding to spatial coordinates. To capture global context information, we take the global context features $\mathbf{g}$ extracted from DINOv2 as input, followed by max pooling and two MLP layers: $\mathbf{T}_\text{global}= \text{MLP}_\text{global}(\text{MaxPool}(\mathbf{g}))$. Following previous work \cite{lhm,yang2024cogvideox}, we use global features to condition Adaptive Layer Normalization (AdaLN), producing separate modulation for geometric tokens and VETs during attention-based fusion.

\subsubsection{Visibility-conditioned Point–Image Attention}
With the geometric and visual tokens constructed, we next introduce a visibility-conditioned point-image attention module to regulate evidence transfer under self-occlusion. The key challenge is that the reliability of visual evidence is inherently visibility-dependent, as occluded points often lack reliable local image evidence. To address this, we first estimate point visibility by a z-buffer test under the posed hand geometry, and complement the local VETs with a global appearance context to support occluded points. Conditioned on the estimated visibility and the local-global visual evidence, our visibility-conditioned point-image cross-attention emphasizes local cues for visible points while shifting attention toward global context for occluded points, yielding occlusion-aware visual context for Gaussian feature construction, as depicted in fig. \ref{fig:transformer}.

Given a projected 3D point $\mathbf{p}_i^{\text{proj}}$ in the camera space with depth $z_i$, we sample the corresponding depth value $z^{\text{buf}}(\mathbf{p}_i^{\text{proj}})$ from the rendered depth map, where $z^{\text{buf}}(\cdot)$ denotes the depth-buffer value of the visible posed hand mesh at the queried image location. We then compute the depth difference and map it to a soft visibility score by sigmoid to obtain a continuous visibility score:
\begin{equation} 
\setlength{\abovedisplayskip}{5pt} 
\setlength{\belowdisplayskip}{5pt}
\begin{gathered} 
\Delta z_i = z^{\text{buf}}(\mathbf{p}_i^{\text{proj}}) - z_i \\ 
p_i^{\text{vis}} = \sigma\!\left({\Delta z_i}/{\tau}\right) 
\end{gathered} 
\end{equation} 
where $\Delta z_i$ is the signed depth residual between the point and the z-buffer depth at $p_i^{\text{proj}}$. The sign encodes whether the point lies in front of or behind the visible surface, while the magnitude measures the depth deviation. $\tau$ controls the softness of the visibility transition. The resulting score $p_i^{\text{vis}} \in [0, 1]$ serves as a geometric proxy for the reliability of local image evidence associated with each 3D point.

To provide occluded points with robust appearance guidance, we introduce a global appearance context as an alternative source of image evidence. By augmenting the local VETs with a global visual token, they form a unified set of image tokens that jointly represent fine-grained local cues and coarse but robust global appearance information. Given this multi-source image evidence, the estimated visibility score naturally serves as a point-wise prior on the reliability of point-image correspondence that governs how each 3D point should select and weight the different sources of visual information. To this end, we incorporate visibility as an additive bias in the point-to-image cross-attention, allowing it to modulate relative attention preference. Concretely, given the soft visibility score $p_i^{\text{vis}}$ for point $i$, we first map it to a signed scalar $s_i = 2p_i^{\text{vis}}-1$, where $s_i \in [-1, 1]$ encodes the directional preference of visual evidence: positive values indicate reliable local cues, while negative values suggest local evidence is likely corrupted by occlusion and global context should be preferred. Based on this formulation, we define a visibility-conditioned attention bias $b_{i,j}$ between point $i$ and image token $j$ as:
\vspace{-5pt}
\begin{equation}
\setlength{\abovedisplayskip}{-5pt} 
\setlength{\belowdisplayskip}{5pt}
b_{i,j} =
\begin{cases}
\;\;\beta_{\text{local}}\, s_i, & j \in \mathcal{J}_{\text{VET}} \\
-\beta_{\text{global}}\, s_i, & j \in \mathcal{J}_{\text{global}}
\end{cases}
\end{equation}
where $\mathcal{J}_{\text{VET}}$ and $\mathcal{J}_{\text{global}}$ denote the index sets of local VETs and the global token, and $\beta_{\text{local}}$ and $\beta_{\text{global}}$ control the bias strength for local and global tokens, respectively. The opposite signs encode a visibility-conditioned preference over local and global image evidence. The visibility-conditioned bias is then added to the attention logits in the point-to-image cross-attention:
\begin{equation}
\begin{gathered}
\tilde{\mathbf{T}}_{3D} = 
\mathrm{softmax}\!\left(
\mathbf{Q}{\mathbf{K}}^\top/\sqrt{d} + \mathbf{B}
\right)\mathbf{V} \\
\mathbf{Q}=W_q\mathbf{T}_{3D}, \quad \tilde{\mathbf{T}}_\text{VET}=\mathbf{T}_\text{VET}+\mathbf{P}_\text{VET}, \quad [\mathbf{K, V}]=W_{kv}\mathbf{T}_{2D}^{\text{full}}.
\end{gathered}
\end{equation}
Here, geometric tokens $\mathbf{T}_{3D}$ serve as the query, and $\mathbf{T}_{2D}^{\text{full}}=[\tilde{\mathbf{T}}_{\text{VET}};\mathbf{T}_{\text{global}}]$ are the key and value. $W_q$ and $W_{kv}$ are learnable projections that map the $\mathbf{T}_{3D}$ to the query $\mathbf{Q}$ and the visual tokens to the key $\mathbf{K}$ and value $\mathbf{V}$, respectively, and $d$ is the dimension of the query and key vectors. $\mathbf{P}_\text{VET}$ encodes the 2D location of VETs. Therefore, $\mathbf{QK}^\top$ reflects the spatial correspondence between each 3D geometric query and the 2D visual tokens, indicating how relevant each visual token is to the queried point. After softmax normalization, these correspondence weights are used to aggregate the value features, allowing each queried point to read visual evidence from the attended visual tokens. $\mathbf{B}$ is the bias matrix formed by $b_{i,j}$, and ${\tilde{\mathbf{T}}_{3D}}$ is the occlusion-aware visual information. Consequently, the resulting point-to-image attention weights allocate more mass to local tokens for visible points ($s_i > 0$) and shift mass towards the global token for occluded points ($s_i < 0$). By injecting visibility-conditioned bias at the logit level, our method effectively modulates evidence reading, encouraging visible points to attend to local visual tokens while allowing occluded points to rely more on global context. The resulting occlusion-aware visual context are then added to the original geometric tokens to form the occlusion-aware Gaussian features.

\subsubsection{Self-Attention Refinement of Occlusion-Aware Gaussian Features}
Although visibility-conditioned cross-attention injects image evidence into geometry tokens, it updates each token independently and therefore cannot explicitly capture dependencies among the Gaussian features. Since the resulting occlusion-aware Gaussian features encode both geometric and occlusion-aware visual context, these two types of information should be further propagated and coordinated across features for coherent reconstruction. Therefore, we apply self-attention to the occlusion-aware Gaussian features, allowing each Gaussian to be exchanged and refined with structural and visual context from others. This inter-feature propagation improves the contextual consistency of Gaussian representations.

\subsection{Feature-on-Mesh Representation}
\label{sec:fom}
Building on the occlusion-aware Gaussian features, we further introduce a more suitable representation carrier for Gaussian deformation under articulated hand animation. Existing approaches \cite{lhm,wu2025sings} typically rely on vertex-based features for Gaussian decoding, implicitly assuming that vertex-level representations are sufficient to capture both appearance and deformation. However, for highly articulated hands, non-rigid deformation often manifests as local stretching and shearing over mesh faces rather than only at vertices. As a result, decoding Gaussians solely from vertex-wise features may fail to adequately reflect local surface deformation over mesh faces, limiting pose-dependent appearance fidelity. To address this limitation, we introduce a Feature-on-Mesh (FoM) representation as a face-level complement to the fused features.

For each mesh triangle $t_i$ with vertices $\{v_{i,1}, v_{i,2}, v_{i,3} \}$, we introduce a set of learnable barycentric coordinate $(u_i^F, v_i^F)$, normalized through a softmax to ensure a valid convex combination, to interpolate vertex-wise features within each face, yielding a face-embedded representation that better captures local surface variation than fixed face sampling. Concretely, given the vertex-wise latent features $\{ \mathbf{F}_{i,1}, \mathbf{F}_{i,2}, \mathbf{F}_{i,3} \}$ obtained from the occlusion-aware Gaussian features, we compute the face-level feature as a barycentric interpolation of the corresponding vertex features:
\begin{equation}
\mathbf{F}^{\text{FoM}}_i
=u_i^F \ast\mathbf{F}_{i,1} + v_i^F \ast\mathbf{F}_{i,2} + (1-u_i^F-v_i^F) \ast \mathbf{F}_{i,3}.
\end{equation}
Each face-level feature is then decoded into Gaussian attributes, including opacity, color, scaling, rotation, and a set of barycentric coordinates $(u_i^G, v_i^G)$ for Gaussian positioning within each triangle. In this way, the decoded Gaussians are explicitly tied to the mesh surface. During animation, we first deform the MANO-HD mesh with LBS and then update Gaussian positions by barycentric interpolation of the posed triangle vertices, ensuring surface-consistent motion. To further adapt Gaussian shape to local non-rigid deformation, we adopt a mesh-driven deformation scheme inspired by \cite{splattingavatar}, where face-level rotations are derived from mesh deformation and used to guide Gaussian transformation (more details are shown in appendix \ref{sec:implementation}). By representing features on mesh faces and coupling Gaussian deformation to mesh deformation, FoM provides a more suitable carrier for pose-dependent Gaussian modeling, leading to consistent appearance reconstruction under hand articulation.

After obtaining the point-wise features and mesh-embedded features, we employ a Gaussian Head to predict 3DGS parameters. For points-wise features, we have:
\begin{equation}
\begin{gathered}
\left\{ \Delta \mathbf{x}_i, \mathbf{r}_i, \mathbf{c}_i, \mathbf{s}_i, \sigma_i \right\}
= \mathrm{MLP}_{\mathrm{GS}}\!\left(\mathbf{T}_{3D}^i\right) \\
\mathbf{p}_i = \mathbf{p}_i + \Delta \mathbf{x}_i,
\quad \forall i \in \{1, \dots, N_{\mathrm{points}}\}
\end{gathered}
\end{equation}
where $\Delta \mathbf{x}_i \in \mathbb{R}^3$ represents residual position offsets from the canonical MANO-HD. For mesh-embedded features, we have:
\begin{equation}
\begin{gathered}
\left\{u_j, v_j, \mathbf{r}_j, \mathbf{c}_j, \mathbf{s}_j, \sigma_j \right\}
= \mathrm{MLP}_{\mathrm{FoM}}\!\left(\mathbf{F}^{\text{FoM}}_j\right) \quad \forall j \in \{1, \dots, N_{\mathrm{FoM}}\} \\
\mathbf{p}_j = u_j \ast \mathbf{p}_{j,1} + v_j \ast\mathbf{p}_{j,2} + (1-u_j-v_j) \ast \mathbf{p}_{j,3}.
\end{gathered}
\end{equation}

Training and one-shot strategies are shown in appendix \ref{sec:implementation}.

\section{Experiments}
\label{sec:exp}

\subsection{Implementation Details}

\noindent \textbf{Learning OASIS.}
We adopt 21 subjects from the InterHand2.6M \cite{moon2020interhand2} training set for pretraining, following \cite{ohta}. During training, we randomly sample a source view image and four target view images from a subject. Our network is trained on two NVIDIA A100 40GB using the Adam optimizer \cite{kingma2014adam} with the learning rates of $4\times10^{-4}$. Loss weights for training objective are set as $\lambda_1=10, \lambda_2=\lambda_3=1, \lambda_4=5$ and $\lambda_p=1, \lambda_s=0.5$.

\noindent \textbf{One-shot Reconstruction.}
We conduct one-shot reconstruction evaluations on the testing set of InterHand2.6M as in \cite{ohta, handavatar} for a fair comparison. For in-the-wild images, we utilize an off-the-shelf estimator \cite{wilor} to predict the MANO parameters and camera poses. We set $\lambda_\mu=20, \lambda_\sigma=1$. The one-shot learning takes 100 steps for color calibration and 400 steps for texture details learning. More details about the experiments are shown in appendix \ref{sec:implementation}.

\newcommand{\best}[1]{\cellcolor{red!10}\textbf{#1}}
\newcommand{\second}[1]{\cellcolor{yellow!30}\underline{#1}}

\begin{table}[ht]
\setlength{\abovecaptionskip}{0.1cm}
\captionsetup{}
\caption{
\textbf{Quantitative results} on InterHand2.6M.
We highlight the \colorbox{red!10}{\textbf{best}} and \colorbox{yellow!30}{\underline{second-best}} results of each metric. LPIPS* = 100 $\times$ LPIPS. $^\S$ means testing without finetuning. 
}
\centering
\label{tab:interhand}

\renewcommand{\arraystretch}{1.3}
\setlength{\aboverulesep}{0.3ex}
\setlength{\belowrulesep}{0.3ex}
\setlength{\tabcolsep}{4pt}{
{\small
\begin{tabular}{>{\centering\arraybackslash}p{4.2mm} l|c|ccc}
\toprule
\multicolumn{2}{c|}{\textbf{Method}} &
\textbf{\#Train} &
\textbf{PSNR$\uparrow$} &
\textbf{SSIM$\uparrow$} &
\textbf{LPIPS*$\downarrow$} \\
\midrule
\midrule

\multirow{3}{*}{\rotatebox{90}{Multi-img}}
& SelfRecon~\cite{jiang2022selfrecon}      & 11,757 & 26.38 & 0.879 & 14.21 \\
& HumanNeRF~\cite{weng2022humannerf}    & 11,757 & 27.64 & 0.884 & 11.45 \\
& HandAvatar~\cite{handavatar}         & 11,757 & 28.23 & 0.894 & 10.35 \\
\midrule

\multirow{7}{*}{\rotatebox{90}{Single-img}}
& HARP~\cite{harp}             & 1 & 19.82 & 0.761 & 22.49 \\
& Handy~\cite{potamias2023handy}           & 1 & 25.56 & 0.794 & 14.98 \\
& HandAvatar~\cite{handavatar} & 1 & 23.79 & 0.820 & 17.78 \\
& OHTA~\cite{ohta}             & 1 & 26.11 & 0.864 & 12.93 \\
& LHM$^\S$~\cite{lhm}              & - & 24.45 & 0.901 & 13.96 \\
& \textbf{Ours$^\S$}  & - &  \second{26.48} & \second{0.950} & \second{11.74} \\
& \textbf{Ours} & 1 & \best{27.38} & \best{0.956} & \best{11.45} \\
\bottomrule
\end{tabular}
}}
\end{table}
\vspace{-10pt}

\begin{figure}[t]
\begin{center}
\setlength{\belowcaptionskip}{0.8cm}
   \includegraphics[width=1\linewidth]{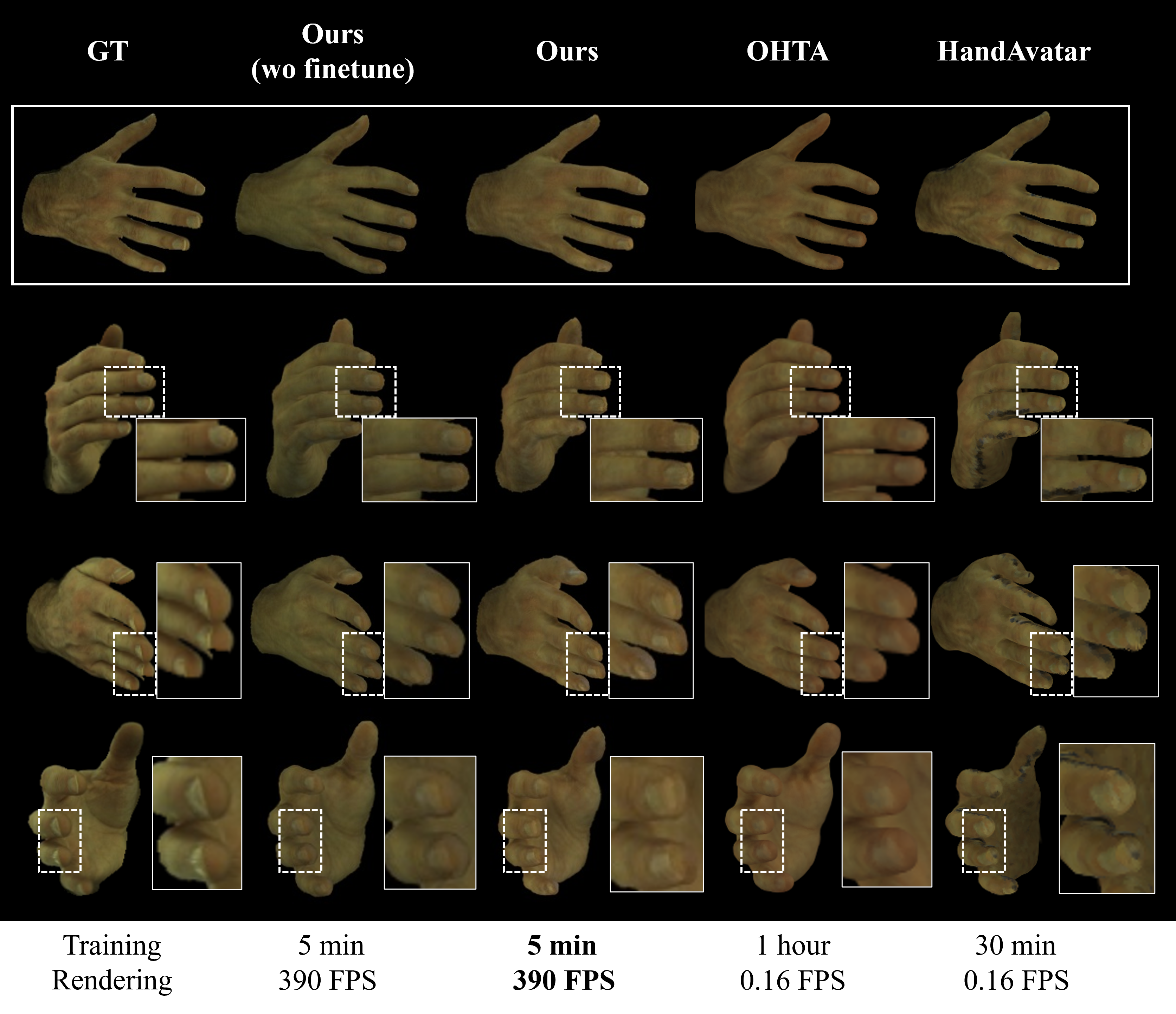}
   \captionsetup{font=small} 
   \captionsetup{skip=4pt}
   \caption {
   Qualitative comparison with state-of-the-art methods on InterHand2.6M \cite{moon2020interhand2}. The white box indicates the input image.
   Frames per second (FPS) are measured on an NVIDIA A100 GPU. 
   } \label{fig:interhand}
\end{center}
\vspace{-30pt}
\end{figure}

\subsection{Evaluation of One-shot Reconstruction}
\noindent \textbf{Quantitative Comparison.}
We quantitatively compare OASIS with previous methods \cite{ohta, handavatar, harp, potamias2023handy, mundra2023livehand, weng2022humannerf, jiang2022selfrecon}. Table \ref{tab:interhand} summarizes the quantitative results of our model against the baselines on the testing set of InterHand2.6M \cite{moon2020interhand2}. We additionally include a no-finetuning variant of our method (denoted as $^\S$), which directly predicts 3D Gaussians from a single input image without one-shot adaptation, to quantify the contribution of the one-shot learning and assess the strength of the prior net. To further contextualize the benefit of our proposed framework, we compare our no-finetuning results with \cite{lhm}, where we train it on the same setting as our prior net training for a fair comparison. Overall, our one-shot reconstruction achieves the superior performance among one-shot methods across all metrics, while the no-finetuning variant still surpasses the baselines, validating the effectiveness of the proposed prior net. Although methods trained with monocular videos \cite{handavatar, weng2022humannerf, jiang2022selfrecon} benefit from substantially stronger input supervision, our one-shot results are the most competitive with them, further demonstrating strong performance under single-image observations. More quantitative comparisons on HanCo dataset \cite{zimmermann2021contrastive} are shown in appendix \ref{sec:hanco_quan}.

\noindent \textbf{Qualitative Comparison.}
As shown in fig. \ref{fig:interhand}, our OASIS demonstrates superior reconstruction quality in InterHand2.6M \cite{moon2020interhand2} compared to baseline models. Although HandAvatar \cite{handavatar} shows impressive results with monocular videos, it fails to generalize well in one-shot scenarios due to the information missing from the single input image. With the prior net and one-shot reconstruction pipeline, OHTA \cite{ohta} achieves comparable visualization results to our OASIS for 10$\times$ more time to converge (around 1 hour). However, it does not perform well on high-frequency texture details, particularly evident in nails. Through our proposed 3D Gaussian-based representation framework, we achieve high-fidelity reconstruction with rapid one-shot learning in approximately 5 minutes. Additionally, our method enables fast rendering at 390 FPS, whereas \cite{ohta, handavatar} require roughly 2500$\times$ more time to render a single image (both operating at around 0.16 FPS). Moreover, our model without finetuning exhibits color discrepancies due to the unseen subject during training, which further validates the effectiveness of the one-shot reconstruction pipeline. Additional qualitative comparisons on the HanCo dataset \cite{zimmermann2021contrastive} are provided in appendix~\ref{sec:hanco_qualitative}.

\begin{figure*}[t]
    \centering
    
    \begin{minipage}[t]{0.3\textwidth}
    \vspace{0pt}
        \centering
        \setlength{\tabcolsep}{4pt}
        \renewcommand{\arraystretch}{1.1}
        \setlength{\aboverulesep}{0.3ex}
        \setlength{\belowrulesep}{0.3ex}
        \begin{tabular}{l l c c c}
            \toprule
            \# & Method & PSNR$\uparrow$ & LPIPS*$\downarrow$ & SSIM$\uparrow$ \\
            \midrule
            a   & Full Model         & \textbf{27.38} & \textbf{11.45} & \textbf{0.956} \\
            \midrule
            b   & w/o VETs           & 26.97 & 12.11 & 0.955 \\
            b.1 & w/o Aggr.          & 27.08 & 11.87 & 0.955 \\
            b.2 & w/o LTEP           & 27.12 & 11.98 & 0.955 \\
            \midrule
            c   & w/o VPIA           & 27.10 & 12.09 & 0.955 \\
            c.1   & w/o bias           & 27.07 & 12.08 & 0.955 \\
             \midrule
            d   & w/o FoM            & 27.13 & 12.35 & 0.955 \\
            \bottomrule
        \end{tabular}
        \vspace{3pt}
        \captionof{table}{Ablation studies of OASIS under the one-shot setting. }
        \label{tab:ablation_prior}
    \end{minipage}
    \hfill
    \begin{minipage}[t]{0.65\textwidth}
    \vspace{0pt}
        \centering
        \captionsetup{skip=4pt}
        \includegraphics[width=\linewidth]{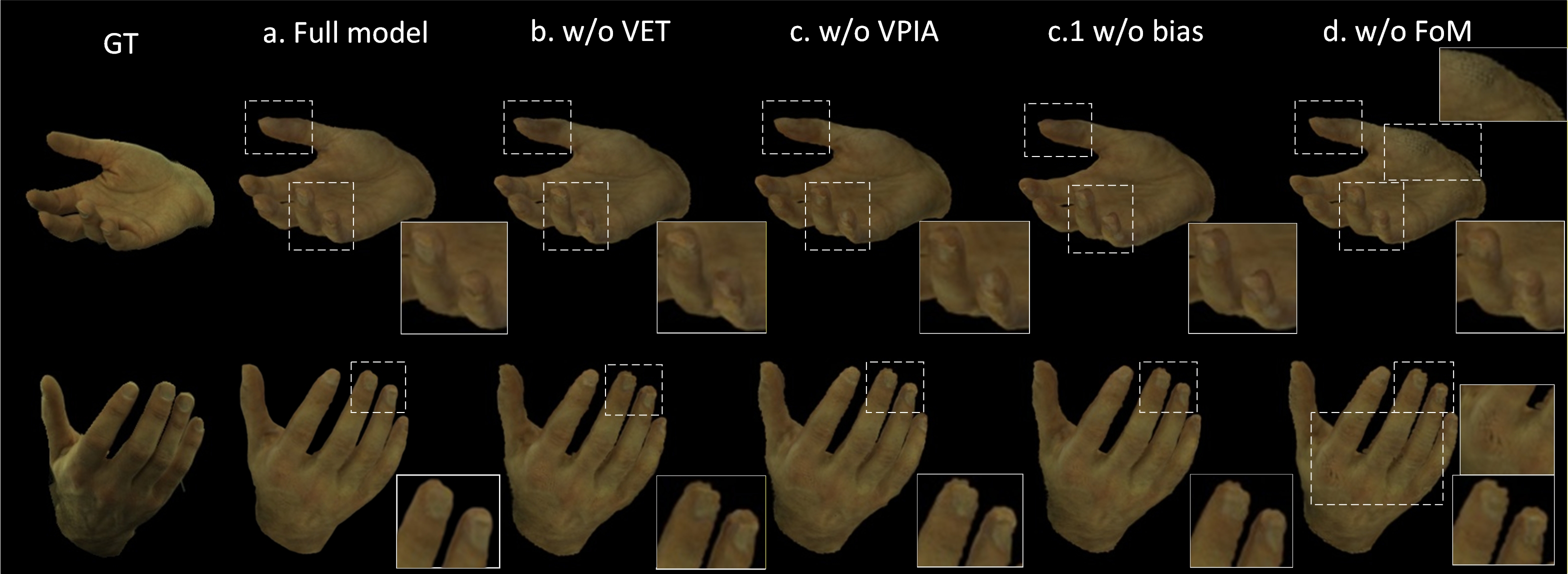}
        \caption{Visual results for ablation studies of OASIS on InterHand2.6M dataset.}
        \label{fig:ablation-mova}
    \end{minipage}
\vspace{-20pt}
\end{figure*}

\subsection{Ablation study}
Tab. \ref{tab:ablation_prior} summarizes the quantitative results of the effectiveness for each component, which are performed on Interhand2.6M dataset. For each ablated variant, we first train the corresponding model with the target component removed, then perform one-shot personalization using the resulting model, and finally evaluate it under the same one-shot setting. Fig. \ref{fig:ablation-mova} presents qualitative results to highlight their contributions. More ablation studies on one-shot reconstruction strategy are shown in appendix \ref{sec:ablation one-shot}.

\noindent \textbf{w/o VET.}
To validate the effectiveness of the proposed geometry-aligned visual evidence tokens construction, we conduct ablation study \#b, which removes the VETs and directly applies image features from the backbone instead, as suggested in \cite{lhm}. Fig. \ref{fig:ablation-mova} b shows that removing VETs leads to blurry local structures and weak fine-grained appearance recovery of fine-grained appearance details, with noticeable distortions in high-frequency regions such as the fingertips and fingernails. In contrast, our full model preserves clearer contours and richer local detail by explicitly aligning image evidence with the corresponding 3D hand geometry through geometry-aligned VETs, which reduces ambiguity in associating observed appearance cues with the correct 3D geometry. This visual improvement is consistent with the better PSNR and LPIPS reported in tab. \ref{tab:ablation_prior} \#b. Additionally, we conduct two further ablation studies on the aggregation and LTEP modules. Specifically, we remove the aggregation step and directly use the projected features to assess the importance of feature aggregation. For the LTEP module, we replace it with a MLP to evaluate its contribution. In both cases, tab. \ref{tab:ablation_prior} \#b.1 and \#b.2 verify that removing the corresponding module leads to degraded performance. The former improves the consolidation of point-aligned local visual cues into more reliable geometry-aligned evidence, while the latter converts the resulting dense evidence into compact and informative visual tokens for subsequent fusion. The performance drop in both cases confirms that both components are necessary for the effectiveness of the complete VETs design.

\noindent \textbf{w/o VPIA.}
For robust image evidence transfer under severe self-occlusion, we introduce a visibility-conditioned point-image attention. To evaluate its benefit, we perform an ablation study \#c to remove VPIA module and instead apply a multimodal transformer in \cite{lhm} to jointly model geometric tokens and VETs. In addition, we further conduct a bias-only ablation, where the visibility bias in VPIA is removed while keeping the remaining attention formulation unchanged. Without the full VPIA mechanism, the reconstruction becomes less reliable in occluded regions, where the partially occluded fingernail area exhibits locally inconsistent structure and texture, as shown in fig. \ref{fig:ablation-mova} c. Removing only the bias leads to noticeable degradation in fig.~\ref{fig:ablation-mova} c.1, where chaotic textures appear in occluded fingertip regions, indicating that the visibility-conditioned bias is crucial for suppressing unreliable local evidence for occluded regions. Our full model alleviates this issue by modulating image evidence transfer according to visibility, allowing visible points to incorporate reliable local image cues while encouraging occluded points to rely more on global information. This advantage is further supported by the better PSNR and LPIPS in tab. \ref{tab:ablation_prior} \#c and \#c.1.

\noindent \textbf{w/o FoM.}
To better support hand deformation, we introduce the FoM representation, which lifts features onto mesh faces to provide a face-level basis for Gaussian modeling. To prove the contribution of the FoM, we ablate this representation in experiment \#d, which only utilizes the fused point features for decoding and animation. As shown in fig. \ref{fig:ablation-mova} d, removing FoM causes noticeable artifacts during pose changes, particularly local holes in regions undergoing surface stretching, where the Gaussian representation fails to maintain sufficient surface coverage. By contrast, our full model preserves continuous hand surfaces and structurally stable local regions by using FoM as a face-level feature carrier that drives the decoded Gaussians to deform consistently with mesh face deformation, yielding coherent appearance across poses without noticeable holes or discontinuities. This visual difference is consistent with the significant drop in LPIPS in tab. \ref{tab:ablation_prior} \#d, suggesting that removing FoM representation weakens the model’s ability to preserve a coherent and stable appearance during non-rigid deformation.

\subsection{Robustness towards Diverse Input and Applications}
To demonstrate the robustness of OASIS, we conduct experiments on diverse input images both quantitatively and qualitatively. The quantitative comparisons are shown in appendix \ref{sec:robust quantitative}. Furthermore, we show more practical applications on text-to-avatar, where the hand avatar is reconstructed by the text prompts, and hand avatar editing. To illustrate the effectiveness of our model, we make comprehensive comparisons with \cite{ohta} on every input image and animation result.In appendix fig. \ref{fig:in-the-wild}, we present single-image 3D hand reconstructions on various in-the-wild images, where the first row for each image shows our reconstruction avatars and the second row shows the results reconstructed from OHTA \cite{ohta}. Notably, OHTA struggles under highly articulated hand poses (e.g., the first row), indicating limited robustness to large pose variations. In contrast, our method preserves richer and more faithful fine-grained appearance details, including palm wrinkles and ring structures. Additionally, we animate the reconstructed avatar for comparison, where OHTA tends to produce texture artifacts along the boundary between observed and unobserved regions, whereas our method maintains consistent textures and smooth transitions. These results demonstrate that our approach generalizes effectively to unseen poses, yielding high-fidelity textures and coherent animation.

Based on our one-shot reconstruction pipeline, we further demonstrate two downstream applications: text-to-avatar and hand avatar editing. For hand avatar editing, we can draw arbitrary content on any single image and then utilize OASIS to reconstruct the target hand avatar, as illustrated in fig. \ref{fig:teaser}. For text-to-avatar, we generate hand images through a hand mask and text prompts following \cite{ohta}, and then reconstruct animatable 3D hand avatars from the generated images. 
The resulting avatars support user-specified animation while preserving the input-view appearance.
As shown in fig.~\ref{fig:app} in appendix, \cite{ohta} tends to smooth out textures under such synthesized images and often fails to reconstruct high-frequency details (e.g., accessories and nails). In comparison, our method preserves finer-grained textures in the input view and maintains consistent appearance during animation, resulting in improved identity and texture fidelity. These results demonstrate that OASIS is more robust to diverse input appearances and can maintain better faithful reconstruction with animation consistency in one-shot settings.

\section{Conclusion}
In this work, we propose a hand-tailored 3DGS model for reconstructing animatable hand avatars from a single image. To address the sparse appearance evidence and severe self-occlusion, we introduce geometry-aligned visual evidence tokens to preserve image-specific appearance cues and and a visibility-conditioned point-image attention module to extract occlusion-aware visual context. By combining this visual context with geometric tokens, we construct occlusion-aware Gaussian features, enabling faithful and robust reconstruction under challenging poses and self-occlusion. Furthermore, our Feature-on-Mesh representation enables Gaussian deformation to better follow local surface deformation, improving reconstruction consistency. Extensive experiments demonstrate that OASIS achieves superior reconstruction quality and efficiency compared with existing approaches, highlighting its effectiveness for animatable hand avatar reconstruction from a single image.

\bibliographystyle{ACM-Reference-Format}
\balance
\bibliography{sigconf}

\clearpage
\appendix
\section*{Appendix}
\setcounter{page}{1}
\section{Preliminary}
\label{preliminary}

\noindent \textbf{3D Gaussian Splatting}~\cite{3dgs} is an explicit 3D representation based on a set of Gaussian primitives, which supports real-time rendering through differentiable rasterization. Specifically, each Gaussian primitive is parameterized by its position center $\mathbf{\mu}$ and a full 3D covariance matrix $\mathbf{\Sigma}$ in a world space, i.e., 
\begin{equation}
\setlength{\abovedisplayskip}{5pt}
\setlength{\belowdisplayskip}{5pt}
\begin{aligned}
G(\mathbf{x})= e^{- \frac{1}{2} (\mathbf{x-\mu})^\top \mathbf{\Sigma}^{-1}\ (\mathbf{x-\mu})},  
\end{aligned}
\label{eqn:gau}
\end{equation}
where $\mathbf{x}$ is any position in 3D space. To guarantee that the covariance matrix $\mathbf{\Sigma}$ is positive semi-definite, it is factorized into a scaling matrix $\mathbf{S}$ and a rotation matrix $\mathbf{R}$, defined as $\mathbf{\Sigma}=\mathbf{R}\mathbf{S}\mathbf{S}^\top\mathbf{R}^\top$. In implementation, each Gaussian is described by a diagonal scaling vector $\mathbf{s}\in\mathbb{R}^3$ and a quaternion vector $\mathbf{r}\in\mathbb{R}^4$, from which a valid covariance matrix can be readily reconstructed. Furthermore, each Gaussian is associated with an opacity value $\alpha \in [0,1)$ and a color vector $\mathbf{c}\in\mathbb{R}^3$. As a result, a set of 3D Gaussians is represented by the parameters $\mathcal{G} = \{\mathbf{\mu, s, r, c}, \alpha\}$. After projecting 3D Gaussians onto the 2D image plane, denoted as $G^{\prime}$, the color of a pixel $p$ is obtained by $\alpha$-blending the $N$ ordered Gaussians that overlap with it:
\begin{equation}
\begin{aligned}
\hat{\mathcal{I}}(p) = \sum_{i=1}^{N} \mathbf{c}_{i} \sigma_{i} \prod_{
j=1}^{i-1}(1-\sigma_j), \quad \sigma_i = \alpha_i G_i^{\prime}(p),
\end{aligned}
\label{eqn:render}
\end{equation}
where $\sigma_i$ is the opacity contribution of the \textit{i}-th Gaussian on the pixel $p$.

\noindent \textbf{MANO} \cite{mano} 
is a pretrained parametric hand model representing hand shape and pose by defining $\mathbf{\theta, \beta}$. It has been widely adopted in prior hand modeling and avatar works \cite{corona2022lisa,handavatar,ohta,harp} to model articulated hand deformation. In this work, we use MANO to animated canonical 3D Gaussians from canonical space to posed space. Specifically, the 3D position and covariance of each 3D Gaussian is translated and rotated by the estimated LBS transformation matrix, 
\begin{equation}
\begin{gathered}
\mathbf{p}^{t}  = \mathbf{G}(\mathbf{J}^{t}, \mathbf{\theta}^{t}) \mathbf{p}^{c} 
  + \mathbf{b}(\mathbf{J}^{t}, \mathbf{\theta}^{t}, \mathbf{\beta}^{t}) \\[4pt]
\mathbf{\Sigma}^{t}  = \mathbf{G}(\mathbf{J}^{t}, \mathbf{\theta}^{t}) 
  \mathbf{\Sigma}^{c} 
  \mathbf{G}(\mathbf{J}^{t}, \mathbf{\theta}^{t})^\top,
\end{gathered}
\label{eqn:smpl}
\end{equation}
where $\mathbf{p}^{t}$, $\mathbf{\Sigma}^{t}$, $\mathbf{p}^{c}$, and $\mathbf{\Sigma}^{c}$ are the position vector and covariance matrix in the posed and canonical space, respectively. $\mathbf{G}(\mathbf{J}^{t}, \mathbf{\theta}^{t}) = \sum_{k=1}^{K} w_k \mathbf{G}_k(\mathbf{J}^{t}, \mathbf{\theta}^{t})$, $\mathbf{b}(\mathbf{J}^{t}, \mathbf{\theta}^{t}, \mathbf{\beta}^{t}) = \sum_{k=1}^{K} w_k \mathbf{b}_k(\mathbf{J}^{t}, \mathbf{\theta}^{t}, \mathbf{\beta}^{t})$ are the rotation matrix and translation vector. $K$ is the joint number, $\mathbf{G}_k(\cdot)$ and $\mathbf{b}_k(\cdot)$ are the transformation matrix and translation vector of joint $k$ respectively, $w_k$ is the LBS weight.

\begin{figure*}[th]
\begin{center}
\setlength{\belowcaptionskip}{1cm}
   \includegraphics[width=1\linewidth]{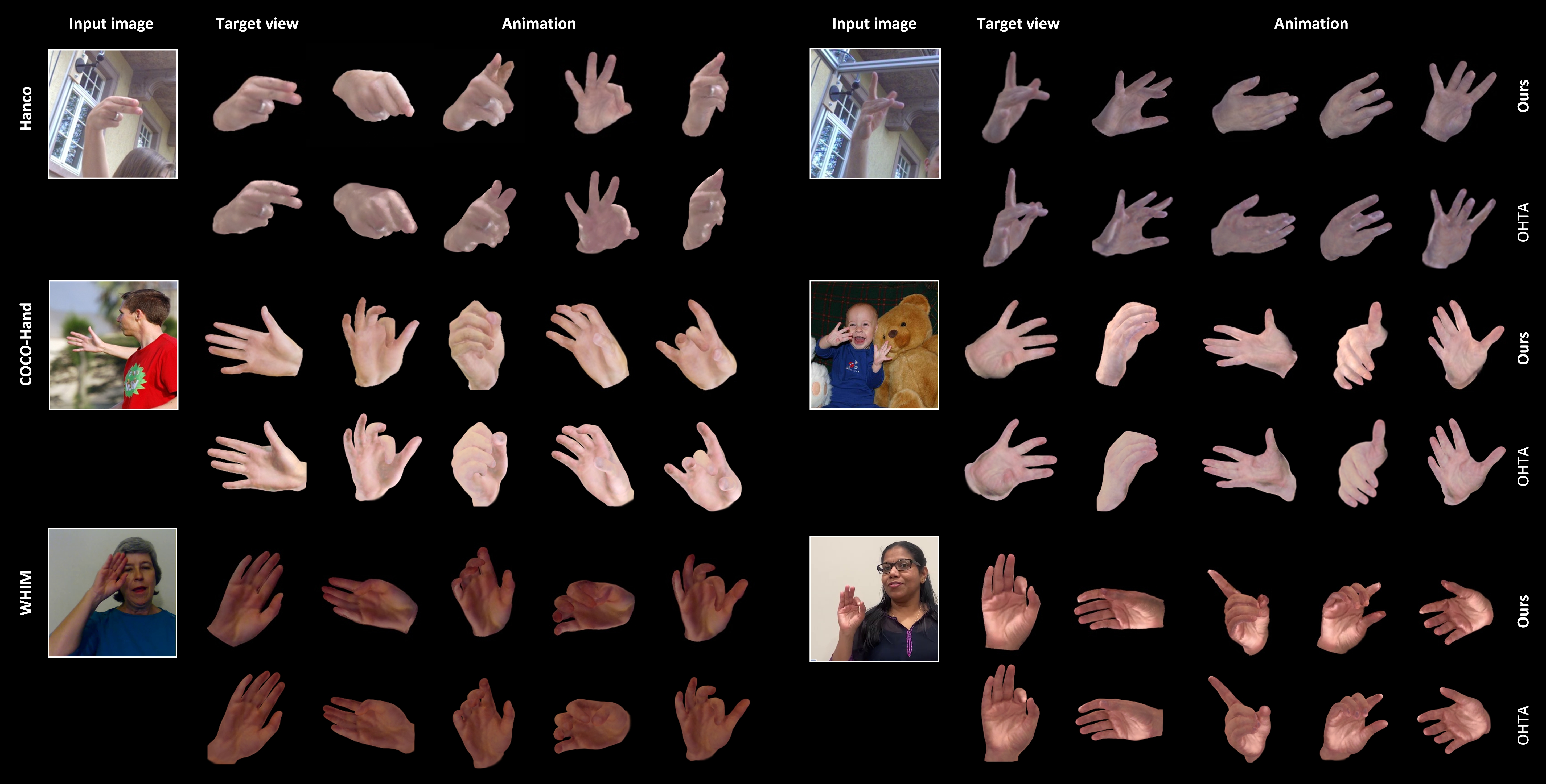}
   \captionsetup{font=small} 
   \caption {
   In the wild visualization from the HanCo \cite{zimmermann2021contrastive} dataset, COCO-Hand \cite{narasimhaswamy2019contextual} dataset, and WHIM \cite{wilor} dataset. 
   } \label{fig:in-the-wild}
\end{center}
\vspace{-25pt}
\end{figure*}

\section{Implementation details of OASIS}
\label{sec:implementation}
\subsection{FoM Representation.} 
The FoM representation organizes features on mesh faces, serving as a face-level bridge between the Gaussian representation and local mesh deformation. However, during animation, simply posing Gaussians with LBS is insufficient to capture the local non-rigid deformation induced by hand articulation. To bridge this gap, we further use mesh-derived  deformation to adapt Gaussian shapes according to local mesh stretching. Specifically, we adopt a mesh-embedded Gaussian deformation scheme to estimate per-vertex local rotations by area-weighted averaging over adjacent triangles, and interpolate them within each face to obtain a face-level rotation. This face-level rotation is composed with the decoded canonical Gaussian rotation $\mathbf{q}_i^{\text{cano}}$ to determine the posed Gaussian orientation $\mathbf{q}_{i}^\text{posed}$: 
\begin{equation}
\begin{gathered}
\mathbf{q}_v = \frac{\sum_{k \in \Omega_v} A_k\,\mathbf{q}_k}{\sum_{k \in \Omega_v} A_k} \\
\delta \mathbf{q}_{i,t} = u_i^G\,\mathbf{q}_{v_{i,1}} + v_i^G\,\mathbf{q}_{v_{i,2}} + (1-u_i^G-v_i^G)\,\mathbf{q}_{v_{i,3}} \\
\mathbf{q}_{i}^\text{posed} = \delta \mathbf{q}_{i,t} \ast \mathbf{q}_i^{\text{cano}} 
\end{gathered}
\end{equation}
where $\Omega_v$ is the neighbor triangles of vertex $v$, $A_k$ and $\mathbf{q}_k$ are the triangle's area and quaternion respectively. Similarly, Gaussian scaling is modulated by the relative area change of each embedded triangle between canonical and posed space: $\mathbf{s}_i^{\text{posed}} =(A_i^{\text{posed}}/A_i^{\text{cano}}) \cdot \mathbf{s}_i^{\text{cano}}$, where $A_i^{\text{cano}}$ and $A_i^{\text{posed}}$ denote the triangle area in canonical and posed space, respectively.

\subsection{Training Strategy}
Given the predicted 3DGS attributes $\mathcal{G}=\{\mathbf{p}, \mathbf{r}, \mathbf{c}, \mathbf{s}, \sigma \}$, representing Gaussian position, rotation, color, scaling, and opacity, respectively, the canonical hand Gaussians are first transformed to target poses by LBS deformation. We then render the deformed Gaussians by differentiable splatting to obtain the rendered image $\hat{\mathcal{I}}$ and the rendered mask $ \hat{\mathcal{M}}$. The view-consistent supervision comprises four components in view space:
\begin{equation} 
\begin{aligned} 
\mathcal{L}_{\textrm{rec}} = \lambda_{1}\mathcal{L}_{\textrm{rgb}} + \lambda_{2}\mathcal{L}_{\textrm{mask}} + \lambda_{3}\mathcal{L}_{\textrm{ssim}} + \lambda_{4}\mathcal{L}_{\textrm{lpips}}.
\end{aligned} 
\end{equation} 

\noindent \textbf{RGB Loss.}
We use an $\mathcal{L}_1$ loss to compute pixel-wise error between the ground truth image $\mathcal{I}_{gt}$ and the rendered image $\hat{\mathcal{I}}$:
\begin{equation}
\label{loss: color}
\begin{aligned}
\mathcal{L}_\text{rgb} =||\hat{\mathcal{I}} - \mathcal{I}_{gt}||_2.
\end{aligned}
\end{equation}

\noindent \textbf{Mask Loss.} 
We apply the mask loss between the ground truth mask $M$ and the accumulated volume density $\hat{\mathcal{M}}$, defined as:
\begin{equation}
\label{loss: mask}
\begin{aligned}
\mathcal{L}_\text{mask} = ||\hat{\mathcal{M}} - M||_2,
\end{aligned}
\end{equation}

\noindent \textbf{SSIM Loss.} 
We also use SSIM to ensure the structural similarity between the rendered image and the ground truth image:
\begin{equation}
\label{loss: ssim}
\begin{aligned}
\mathcal{L}_\text{ssim} = \text{SSIM}(\hat{\mathcal{I}}, \mathcal{I}_{gt}),
\end{aligned}
\end{equation}

\noindent \textbf{LPIPS Loss.}
To ensure the quality of the rendered image, we utilize the LPIPS perceptual loss, i.e.,
\begin{equation}
\label{loss: lpips}
\begin{aligned}
\mathcal{L}_\text{lpips} = \text{LPIPS}(\hat{\mathcal{I}}, \mathcal{I}_{gt}),
\end{aligned}
\end{equation}

In addition to the reconstruction loss, we introduce additional regularization terms to enforce geometric coherence in canonical space, as suggested in \cite{lhm}. To maintain hand surface plausibility, we encourage Gaussian positions to be close to their initialized locations:
\begin{equation} 
\begin{aligned} 
\mathcal{L}_{\text{position}} =
\frac{1}{N_{\text{points}}}
\sum\nolimits_{i=1}^{N_{\text{points}}}
\max\left(\lVert \Delta p_i \rVert_2 - d, 0\right)
\end{aligned} 
\end{equation} 
where $d$ is an empirically determined threshold. To penalize excessive thin and long Gaussian primitives, we apply a scaling regularization:
\begin{equation}
\label{eq:scaling}
\ell_i = 
\max\left(
\frac{s_i^{\max}}{s_i^{\min}} - r,\, 0
\right), \quad
\mathcal{L}_{\text{scaling}}=\frac{1}{N_\text{point}}
\sum\nolimits_{i=1}^{N_\text{point}} \ell_i
\end{equation}
where $s_i \in \mathbb{R}^3$ is the Gaussian's scaling, and $s_i^{\max}, s_i^{\max}$ are the maximum and minimum scaling values respectively. $r$ is an empirically determined threshold. Therefore, the total loss for training objective is formulated as:
\begin{equation} 
\begin{aligned} 
\mathcal{L}_\text{total} = \mathcal{L}_{\textrm{rec}} + \lambda_{\textrm{p}}\mathcal{L}_\text{position} + \lambda_{\textrm{s}}\mathcal{L}_\text{scaling}.
\end{aligned} 
\end{equation}

\subsection{One-shot Hand Avatar Reconstruction}
In the stage of one-shot hand avatar reconstruction, 
we first utilize an off-the-shelf hand pose estimator \cite{wilor} to get the predicted hand shape parameter $\mathbf{\beta}$, pose parameter $\mathbf{\theta}$ and camera pose. Inspired by previous work \cite{ohta}, we optimize to render an image $\hat{\mathcal{I}}$ that is similar to the target subject of the input image by introducing per-channel color calibration coefficients. Specifically, we introduce $\{ \mathbf{w},\mathbf{b} \} \in \mathbb{R}^{1\times3}$ to modulate the Gaussian appearance by $\tilde{\mathbf{c}} = \mathbf{w}\ast\mathbf{c}+\mathbf{b}$. We only optimize these color-related global parameters under masked photometric and color-statistics supervision, while freezing geometry and texture-related modules to preserve the pretrained appearance prior. For supervision, we compute the masked mean and standard deviation of an image, and enforce the rendered image to match the ground-truth image by minimizing the difference between their masked statistics:
\begin{equation}
\begin{aligned}
\mu(I;M) = \frac{\sum I \odot M}{\sum M} \quad
\sigma(I;M) = \sqrt{\frac{\sum (I - \mu(I;M))^2 \odot M}{\sum M} + \epsilon} \\
\mathcal{L}_{\text{calib}}
=
\lambda_1 \mathcal{L}_\text{rgb}
+
\lambda_\mu \left\|
\mu(\hat{\mathcal{I}}) - \mu(\mathcal{I}_{gt})
\right\|_1
+
\lambda_\sigma \left\|
\sigma(\hat{\mathcal{I}}) - \sigma(\mathcal{I}_{gt})
\right\|_1 \\
\end{aligned}
\end{equation}
After the color calibration stage, we train a LoRA (Low-Rank Adaptation) \cite{hu2022lora} module for Gaussian Head to capture the details of the target identity from the input image while keeping others frozen. To avoid overfitting the target view, we also perform view regularization. Specifically, we constrain the finetuning results of some reference views with different poses $\{ \mathbf{R}_i\}^{N^{r}}_{i=1}$ to be close to the rendering results with the same poses at the end of the color calibration stage. 
\begin{equation} 
\begin{aligned} 
\mathcal{L}_\text{img} = \mathcal{L}_\text{rec}(\mathcal{I}, \hat{\mathcal{I}}) + 
\lambda_\text{ref} \sum\nolimits_{i=1}^{N^{r}}
\left(\mathcal{L}_{rec}(\mathcal{I}_{i}(\mathbf{R}_{i}), \hat{\mathcal{I}}(\mathbf{R}_{i}))\right)
\end{aligned} 
\end{equation} 
where $N^r$ is the number of the generated reference views. To prevent Gaussians from excessive anisotropy, we also perform the scaling regularization in Eq. \ref{eq:scaling}.
\begin{equation} 
\begin{aligned} 
\mathcal{L}_\text{tex} = \mathcal{L}_\text{img} + \lambda_{s} \mathcal{L}_\text{scaling}
\end{aligned} 
\end{equation}

\begin{figure*}[t]
\begin{center}
\setlength{\belowcaptionskip}{1cm}
   \includegraphics[width=1\linewidth]{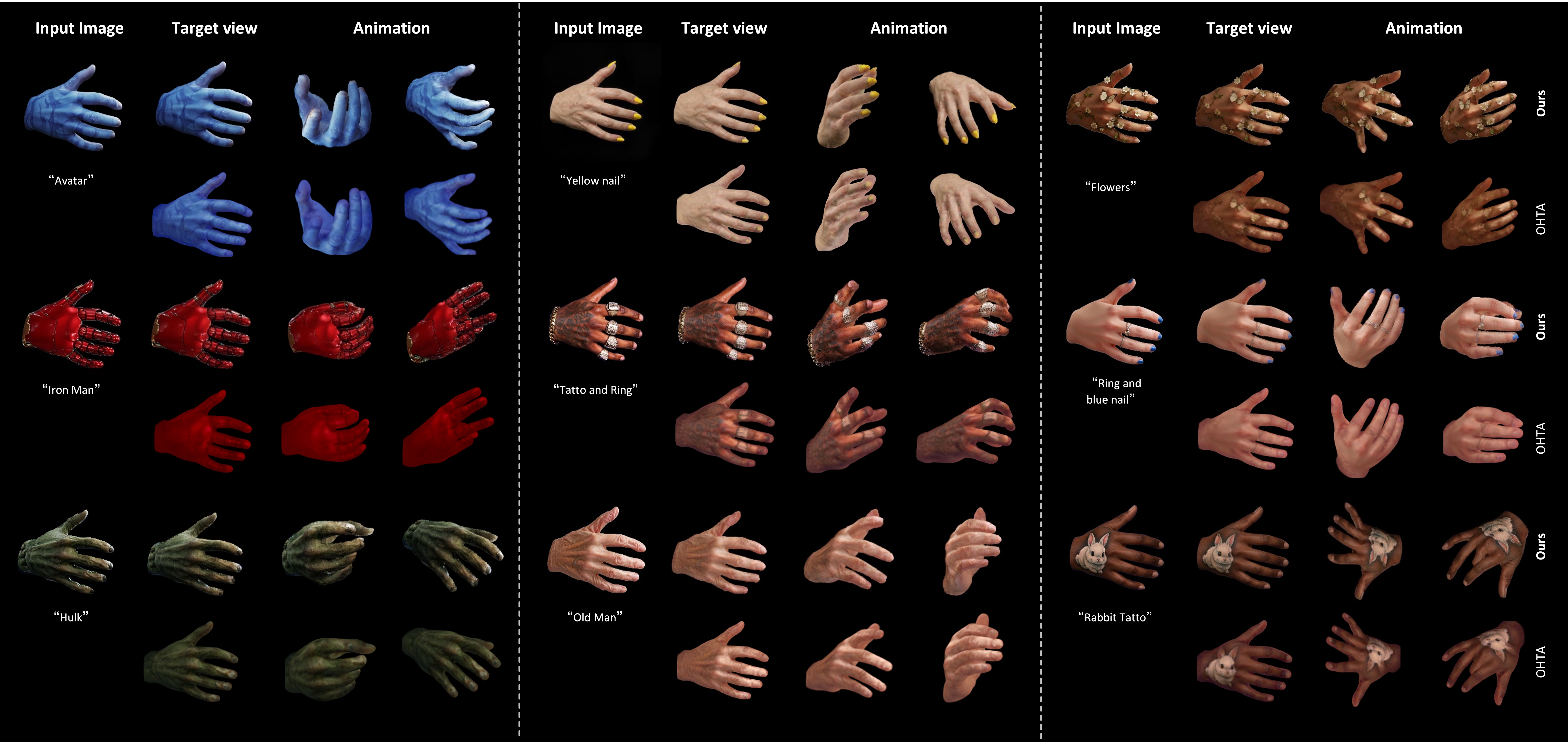}
   \captionsetup{font=small} 
   \caption {
   Qualitative comparison with \cite{ohta} on text-to-avatar application. The first column is the input reference images, and the rendered target views are displayed on the right side, respectively. The following two visual results are obtained by animating the reconstructed 3D hand avatar to different poses.
   } \label{fig:app}
\end{center}
\vspace{-25pt}
\end{figure*}

\section{Experiments}

\subsection{Metrics}
Following prior works \cite{handavatar, ohta}, we use Learned Perceptual Similarity (LPIPS) \cite{lpips}, Structural Similarity Index Measure (SSIM) \cite{ssim}, and Peak Signal-to-Noise Ratio (PSNR) \cite{psnr} as reconstruction quality metrics to measure image similarity between ground truth images and rendered images.

\subsection{Dataset Details}
\noindent \textbf{InterHand2.6M \cite{moon2020interhand2}.} 
For the full model pretraining, we follow \cite{ohta} to use the subject `train/Capture0', `train/Capture1', `train/Capture2', `train/Capture3', `train/Capture5', `train/Capture6', `train/Capture7', `train/Capture8', `train/Capture9', `train/Capture10', `train/Capture11', `train/Capture12', `train/Capture13', `train/Capture14', `train/Capture15', `train/Capture16', `train/Capture20', `train/Capture22', `train/Capture23', `train/Capture24', `train/Capture25' for training. We exclude the pose sequences `0000\_neutral\_relaxed', `0009\_thumbtucknormal', `0019\_alligator\_closed', `0029\_indextip', `0039\_fingerspreadrigid', `0048\_index\_point', `0058\_middlefinger' to ensure a consistent experimental setting with \cite{ohta}. For evaluation of one-shot hand avatars, we follow \cite{handavatar} to use sequence `test/Capture0/ROM03\_RT\_No\_Occlusion' with fixed skip steps. For each frame, we follow \cite{handavatar} and use the annotated detection box as the ground-truth hand region. The box is first adjusted to a square and enlarged by a factor of 1.3, after which the hand region is cropped and resized to a resolution of 256$\times$256.

\noindent \textbf{HanCo \cite{zimmermann2021contrastive}.} 
For quantitative experiments on HanCo dataset, we use sequence `0154' with the cameras 0,3,5,6,7. We apply the provided MANO annotations, camera poses and hand masks of the dataset for one-shot reconstruction.

\noindent \textbf{In-the-wild Data.} 
For in-the-wild evaluation, we use whole-body images from COCO-Hand \cite{narasimhaswamy2019contextual} and WHIM \cite{wilor}. For each input image, we first use WiLoR \cite{wilor} to detect the hand and determine its handedness. We then adopt the right hand for pose estimation, where \cite{wilor} predicts the camera parameters and MANO annotations of the input images. We further use the corresponding pose estimation results to generate hand masks.

\subsection{Quantitative Comparison.} 
\label{sec:hanco_quan}
To show the robustness of different subjects from the training set, we also perform the quantitative comparison on the HanCo dataset \cite{zimmermann2021contrastive} with \cite{ohta, handavatar}, as they show robust one-shot reconstruction on InterHand2.6M. As shown in tab. \ref{tab:hanco}, our method outperform previous works consistently in all metrics. 

\begin{figure}[t]
\begin{center}
\setlength{\belowcaptionskip}{1cm}
   \includegraphics[width=1\linewidth]{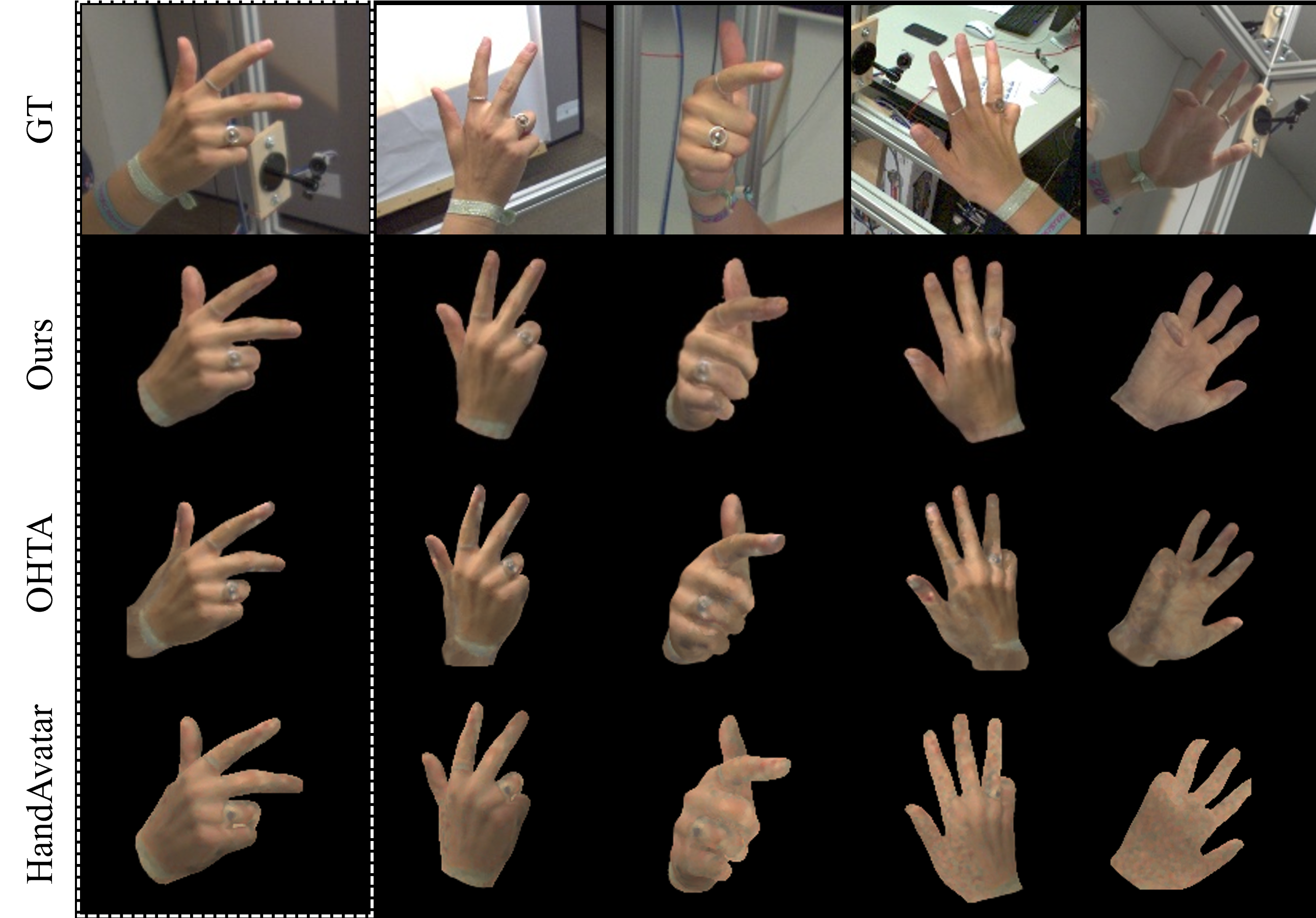}
   \captionsetup{font=small} 
    \captionsetup{skip=4pt}
   \caption {
   Qualitative comparison on HanCo \cite{zimmermann2021contrastive}. The white box indicates the input view.
   } \label{fig:hanco}
\end{center}
\vspace{-20pt}
\end{figure}

\begin{table}
  \centering
  \begin{tabular}{lccc}
    \toprule
    Method & PSNR$\uparrow$ & LPIPS*$\downarrow$ & SSIM$\uparrow$ \\
    \midrule
    HandAvatar~\cite{handavatar}     & 24.68 & 8.25 & 0.896 \\
    OHTA~\cite{ohta}  & 25.24 & 7.18 & 0.923 \\
    Ours                               & \textbf{25.86} & \textbf{6.76} & \textbf{0.974} \\
    \bottomrule
  \end{tabular}
  \caption{Evaluation results on HanCo.}
  \label{tab:hanco}
\end{table}

\begin{figure*}[th]
\begin{center}
\setlength{\belowcaptionskip}{1cm}
   \includegraphics[width=1\linewidth]{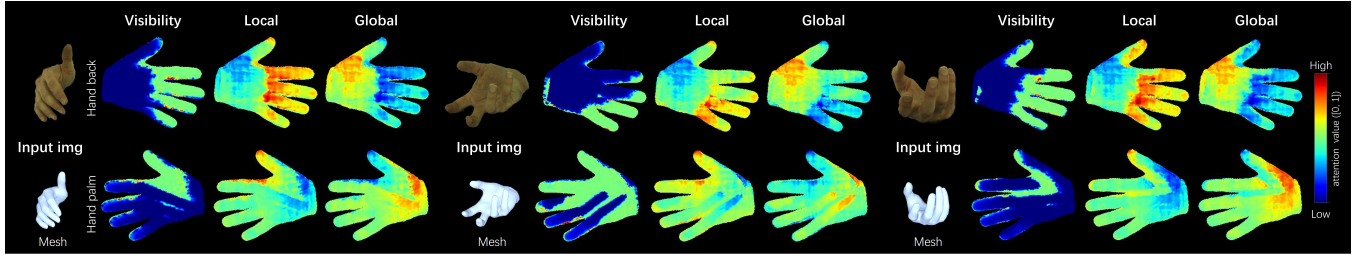}
   \captionsetup{font=small} 
   \caption {
    Points visibility and VPIA attention visualization.
   } \label{fig:attn}
\end{center}
\vspace{-20pt}
\end{figure*}

\subsection{Qualitative Comparison}
\label{sec:hanco_qualitative}
With respect to the qualitative comparison results on the HanCo dataset, fig. \ref{fig:hanco} highlights that our method maintains more consistent appearance across diverse hand poses and viewpoints. In contrast, HandAvatar~\cite{handavatar} fails to recover plausible textures for unseen regions due to the incomplete information provided by single-view conditioning. OHTA~\cite{ohta} tends to produce noisy texture and unstable boundary transitions, especially around the occluded regions. By comparison, our method produces cleaner textures and smoother transitions under large viewpoint variations, leading to more coherent and visually plausible reconstructions.

Qualitative comparisons on in-the-wild images with OHTA \cite{ohta} are shown in fig. \ref{fig:in-the-wild}. We further demonstrate the strong versatility in texture editing and text-to-avatar generation in fig. \ref{fig:teaser} and \ref{fig:app}. For texture editing, we first edit the input image and obtain the corresponding edit mask. During one-shot reconstruction, we exclude the edited region when performing global color calibration, since the edited content may introduce colors that differ significantly from those of the hand and thus bias the global color statistics. We then run the texture learning stage for 1000 steps to adapt the reconstructed texture to the edited appearance and recover its high-frequency details. For text-to-avatar generation, we use Nano Banana with depth maps and text prompts as input for image generation. Subsequently, we perform one-shot reconstruction from these images with 1000 steps to better capture the complex details.

\subsection{Quantitative Comparison for Robustness}
\label{sec:robust quantitative}
\begin{figure}[h]
\begin{center}
\setlength{\belowcaptionskip}{1cm}
   \includegraphics[width=1\linewidth]{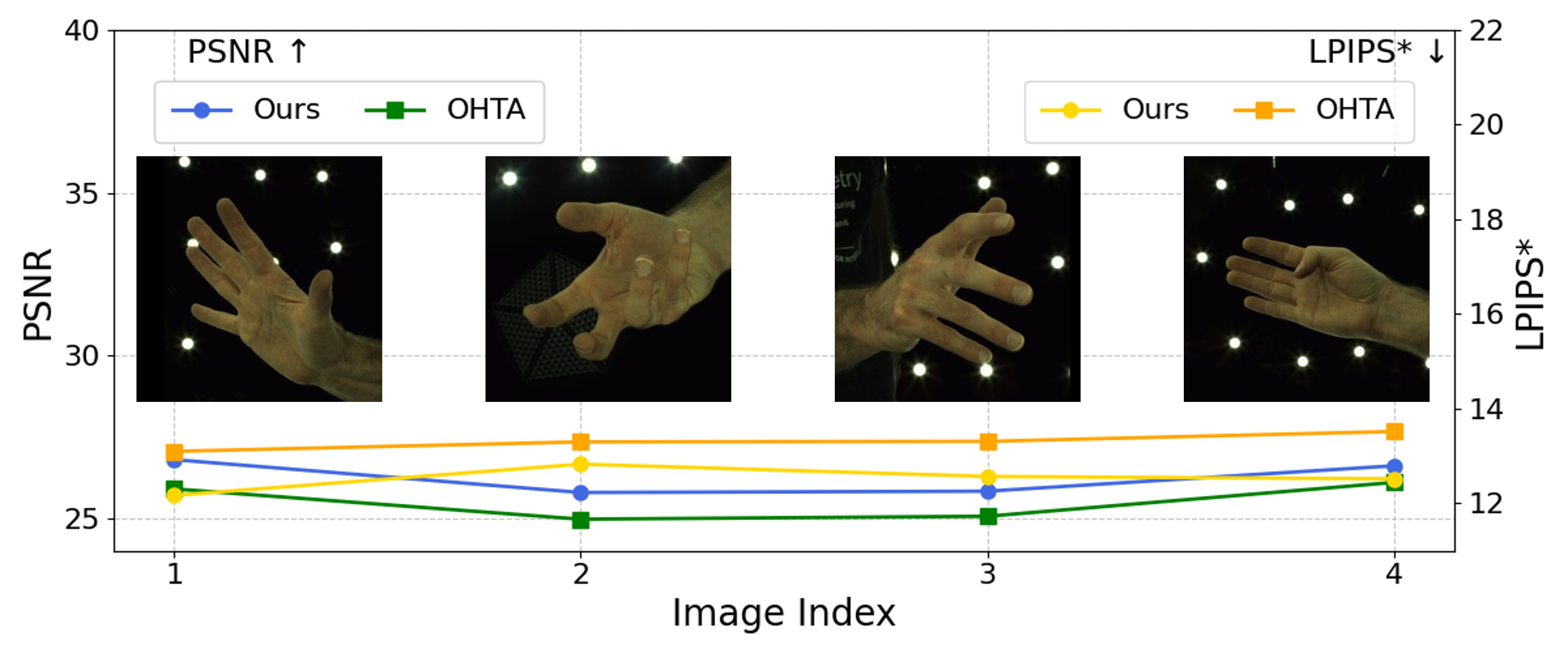}
   \captionsetup{font=small} 
   \caption {
   Robustness towards different input images. The images above the figure are the corresponding input images. 
   } \label{fig:robust}
\end{center}
\vspace{-25pt}
\end{figure}

Fig.~\ref{fig:robust} shows a per-image quantitative comparison between our method and OHTA~\cite{ohta} on different input images from InterHand2.6M. Our method achieves consistently better reconstruction quality, yielding higher PSNR and lower LPIPS across diverse examples. These stable improvements over varying hand poses and viewpoints demonstrate the robustness of the proposed approach.

\begin{figure}[h]
\begin{center}
\setlength{\belowcaptionskip}{1cm}
   \includegraphics[width=1\linewidth]{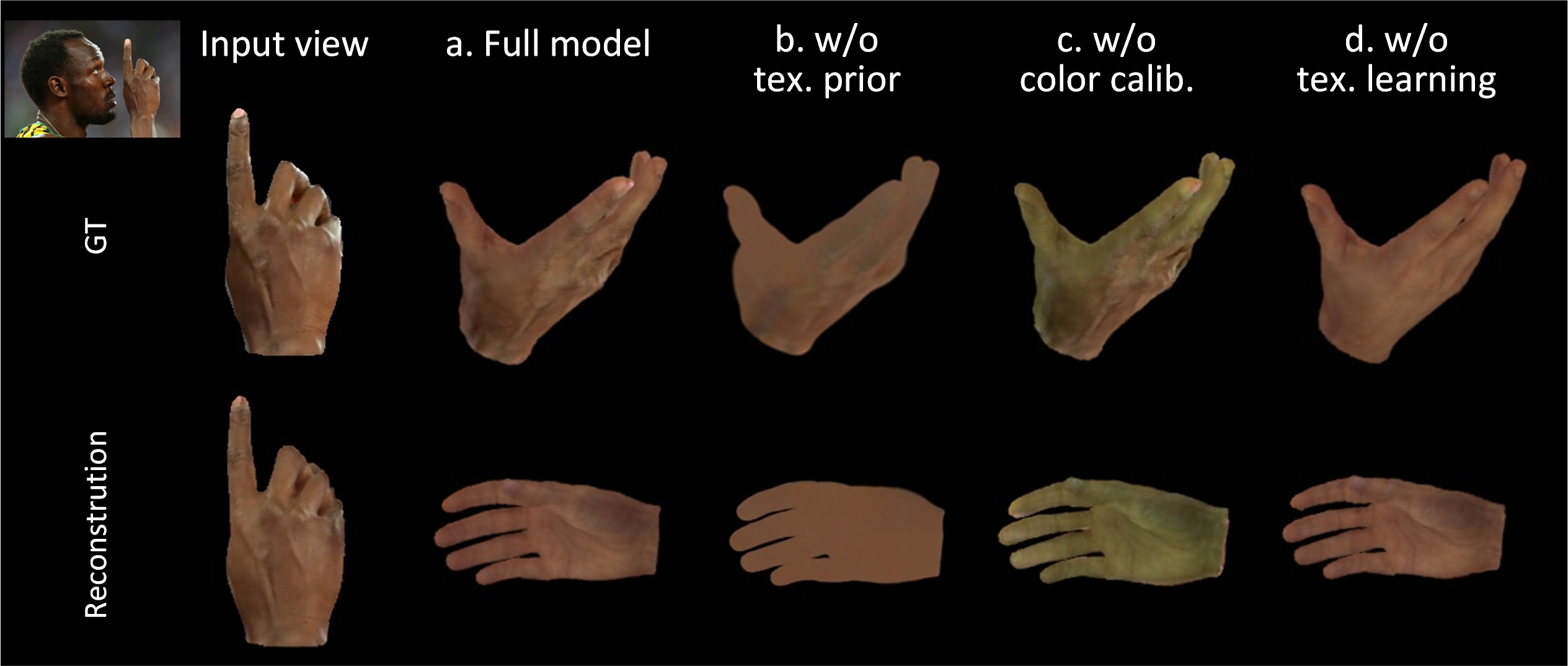}
   \captionsetup{font=small} 
   \caption {
   Visual results for ablation study with in-the-wild input. 
   } \label{fig:ablation}
\end{center}
\end{figure}

\subsection{Ablation Studies}
\noindent \textbf{Evidence routing of VPIA. } 
We visualize the estimated visibility, a point-wise visibility probability obtained by sigmoid normalization to the z-buffer depth residuals, and the attention map in fig.~\ref{fig:attn}. Visible regions in the input view generally show higher local attention, while unseen or self-occluded regions show higher global attention. This confirms that OASIS uses local image evidence for visible regions and shifts to global context for occluded regions, which is consistent with the design of VPIA.

\begin{table}[th]
\centering
\begin{tabular}{c l c c c}
\toprule
\# & Method & PSNR$\uparrow$ & LPIPS*$\downarrow$ & SSIM$\uparrow$ \\
\midrule
a & Full Model & \textbf{27.38} & \textbf{11.45} & \textbf{0.956} \\
\midrule
b & w/o Tex. prior & 25.88 & 16.24 & 0.912 \\
\midrule
c & w/o Color Calib. & 26.59 & 11.98 & 0.951 \\
d & w/o Tex. Learning & 27.02 & 12.18 & 0.953 \\
\midrule
e & w/o Regularization & 27.08 & 12.28 & 0.954 \\
f & w/o LoRA Finetune & 27.11 & 12.06 & 0.954 \\
\bottomrule
\end{tabular}
\vspace{5pt}
\caption{Ablation of one-shot reconstruction on InterHand2.6M. Tex. is short for texture, and calib. is short for calibration.}
\label{tab:ablation_one_shot}
\end{table}

\label{sec:ablation one-shot}
\noindent \textbf{One-shot Reconstruction.}
To validate each component's effectiveness, we perform ablation studies under the one-shot setting. The quantitative results are summarized in tab. \ref{tab:ablation_one_shot} on the InterHand2.6M dataset, and the qualitative results are performed on in-the-wild images, displayed in fig. \ref{fig:ablation}. As shown in tab. \ref{tab:ablation_one_shot} \#b, removing texture priors and instead merely performing the one-shot strategy under the single-view setting leads to consistently worse reconstruction quality. We attribute this degradation to the fact that one-shot supervision is highly limited and under-constrained; without priors knowledge, the model lacks a strong initialization for unseen appearances (as shown in fig. \ref{fig:ablation} b), while full tuning substantially enlarges the optimization space and makes target fitting less stable. Without color calibration, the model suffers from the skin-color bias of the training data, which leads to a noticeable drop in PSNR as shown in tab. \ref{fig:ablation} \#c. As seen in fig. \ref{fig:ablation} c, visible regions can still be learned from the input image, while unseen regions are more strongly determined by the learned prior, retaining clear color discrepancies. This highlights the role of color calibration in reducing the training-domain color bias carried by the prior, which is essential for maintaining coherent appearance reconstruction across the whole hand. Omitting texture learning leads to a significant deterioration in LPIPS, as reported in tab. \ref{tab:ablation_one_shot} \#d, indicating a clear loss in perceptual fidelity. Qualitatively, fig.~\ref{fig:ablation} d shows that the model can only rely on the prior network to preserve a coarse texture structure, while failing to reconstruct high-frequency details. This confirms that texture learning is crucial for injecting fine-grained image-specific cues beyond the generic prior, thereby enabling faithful one-shot appearance reconstruction.

\section{Limitations and Failure Cases}
\begin{figure}[h]
\begin{center}
\setlength{\belowcaptionskip}{1cm}
   \includegraphics[width=1\linewidth]{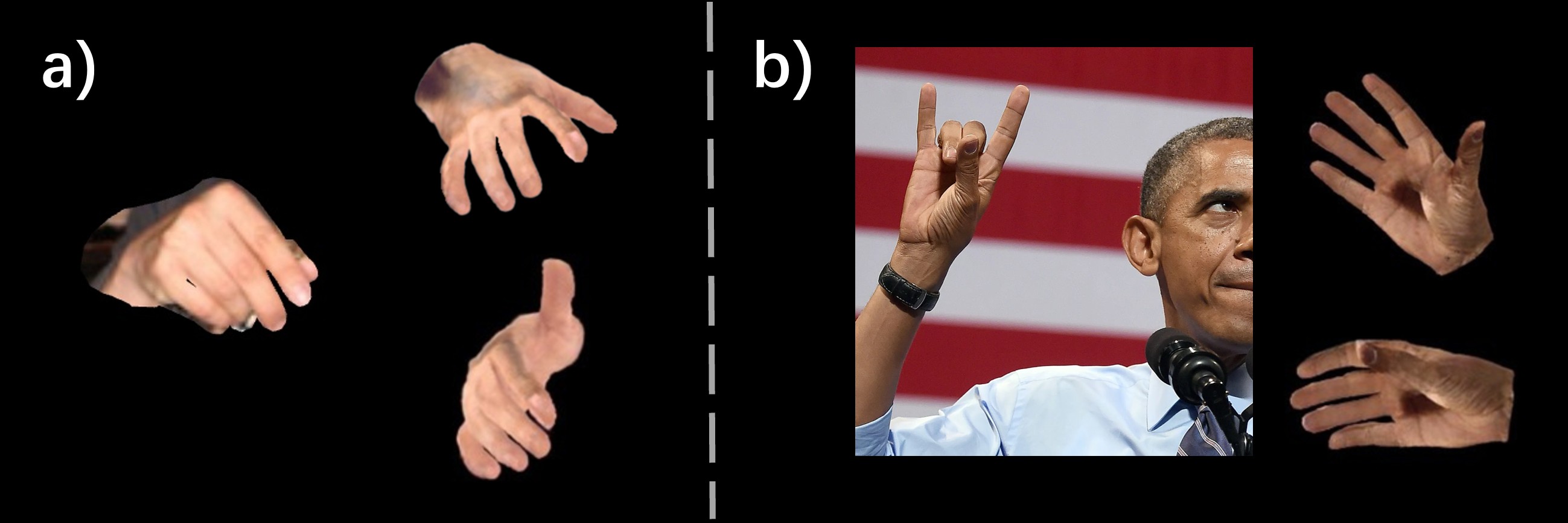}
   \captionsetup{font=small} 
   \caption {
   Failure cases. 
   } \label{fig:limit}
\end{center}
\end{figure}

(a) \textit{Dependency on hand estimators.} 
OASIS depends on off-the-shelf estimators \cite{wilor} for pose, camera, handedness, and masks. Poor estimates, especially for challenging poses or extreme viewpoints, can lead to inferior texture modeling since one-shot personalization is optimized from an unreliable geometric initialization and observation. Although the learned hand prior provides robustness to moderate estimation errors, OASIS cannot fully handle highly inaccurate pose and camera estimates, as shown in fig.~\ref{fig:limit} (a).

\noindent (b) \textit{Shadow/albedo ambiguity.} OASIS reconstructs observed appearance from single images and does not explicitly perform intrinsic decomposition; therefore, strong cast shadows may be baked into the texture (see fig.~\ref{fig:limit} (b)). This is a fundamental single-image inverse-rendering ambiguity, since material, geometry, and lighting are entangled. We will explore integrating material-aware decomposition into OASIS to separate illumination-dependent effects in future work.

\end{document}